\documentclass[smallcondensed]{svjour3}       
\usepackage{helvet} 
\usepackage{courier}  
\usepackage[hyphens]{url}  
\usepackage{graphicx} 
\usepackage[square,sort,comma,numbers]{natbib}  
\usepackage{caption} 
\usepackage{amsmath,amssymb}
\usepackage{bbm}
\usepackage{subfig}
\usepackage{booktabs}
\usepackage{wrapfig}
\usepackage{enumitem}
\usepackage{hyperref}

\usepackage{amsmath,amsfonts,bm}

\def\eqref#1{equation~\ref{#1}}

\def\1{\bm{1}}

\def\ve{{\bm{e}}}
\def\vf{{\bm{f}}}
\def\vg{{\bm{g}}}

\def\vx{{\bm{x}}}
\def\vy{{\bm{y}}}

\def\mI{{\bm{I}}}

\DeclareMathAlphabet{\mathsfit}{\encodingdefault}{\sfdefault}{m}{sl}
\SetMathAlphabet{\mathsfit}{bold}{\encodingdefault}{\sfdefault}{bx}{n}

\def\sR{{\mathbb{R}}}

\newcommand{\E}{\mathbb{E}}
\newcommand{\R}{\mathbb{R}}

\DeclareMathOperator*{\argmin}{arg\,min}

\usepackage[linesnumbered,ruled]{algorithm2e}
\IncMargin{1em}

\SetCommentSty{mycommfont}
\usepackage[utf8]{inputenc}

\usepackage{xcolor}
\newcommand{\lxh}[1]{#1}

\authorrunning{ }
\titlerunning{Interpretable Deep Learning}

\begin{document}
\title{Interpretable Deep Learning: Interpretation, Interpretability, Trustworthiness, and Beyond}

\author{
Xuhong Li\footnotemark[5], 
Haoyi Xiong\footnotemark[5],
Xingjian Li\footnotemark[5],
Xuanyu Wu\footnotemark[6],
Xiao Zhang\footnotemark[7],
Ji Liu\footnotemark[5],
Jiang Bian\footnotemark[5],
Dejing Dou\footnotemark[1]\footnotemark[5]\footnotemark[8]
}

\date{Received: date / Accepted: date}

\institute{
\footnotemark[5] Baidu Research, Baidu Inc., Beijing, China\\
            \email{\{lixuhong, xionghaoyi, lixingjian, liuji04, bianjiang03, doudejing\}@baidu.com}\\
\footnotemark[6] School of Engineering and Applied Science, University of Pennsylvania, Philadelphia, PA \\
            \email{xuanyuwu@seas.upenn.edu} \\
\footnotemark[7] Department of Electronics and Information Engineering, Tsinghua University, Beijing, China \\
            \email{xzhang19@mails.tsinghua.edu.cn} \\
\footnotemark[8] Computer and Information Science Department, University of Oregon, Eugene, OR. \\
\email{dou@cs.uoregon.edu}\\
\footnotemark[1] Correspondence to Dejing Dou via doudejing@baidu.com and dou@cs.uoregon.edu.
}

\maketitle

\begin{abstract}

Deep neural networks have been well-known for their superb handling of various machine learning and artificial intelligence tasks. 
However, due to their over-parameterized black-box nature, it is often difficult to understand the prediction results of deep models.
In recent years, many interpretation tools have been proposed to explain or reveal how deep models make decisions.
In this paper, we review this line of research and try to make a comprehensive survey.
Specifically, we first introduce and clarify two basic concepts---interpretations and interpretability---that people usually get confused about.
To address the research efforts in interpretations, we elaborate the designs of a number of interpretation algorithms, from different perspectives, by proposing a new taxonomy. 
Then, to understand the interpretation results, we also survey the performance metrics for evaluating interpretation algorithms. 
Further, we summarize the current works in evaluating models' interpretability using \emph{``trustworthy''} interpretation algorithms. 
Finally, we review and discuss the connections between deep models' interpretations and other factors, such as adversarial robustness and learning from interpretations, and we introduce several open-source libraries for interpretation algorithms and evaluation approaches.

\end{abstract}

\section{Introduction}

Deep learning models~\cite{DBLP:journals/nature/LeCunBH15} have achieved remarkable performance in a variety of tasks, from visual recognition, natural language processing, reinforcement learning to recommendation systems, where deep models have produced results comparable to and in some cases superior to human experts.
Due to their nature of over-parameterization (involving more than millions of parameters and stacked with more than hundreds of layers), it is often difficult to understand the prediction results of deep models~\cite{doshi2017towards}.
Explaining\footnote{The subtle differences among \textit{interpretation}, \textit{explanation}, and \textit{attribution} are not considered in this paper, and we use them interchangeably.} their behaviors remains challenging because of their hierarchical non-linearity in a black-box fashion.
The lack of interpretability raises a severe issue about the trust of deep models in high-stakes prediction applications, such as autonomous driving, healthcare, criminal justice, and financial services~\cite{carvalho2019machine}.
While many interpretation tools have been proposed to explain or reveal the ways that deep models make decisions, nonetheless, either from a scientific view or a social aspect, explaining the behaviors of deep models is still in progress.
In this paper, instead of focusing on the social impacts, regulations, and laws related to deep model interpretations, we would like to focus on the research field by clarifying the research objectives and reviewing the methods proposed.

\begin{figure}
    \centering
    \includegraphics[width=0.95\linewidth]{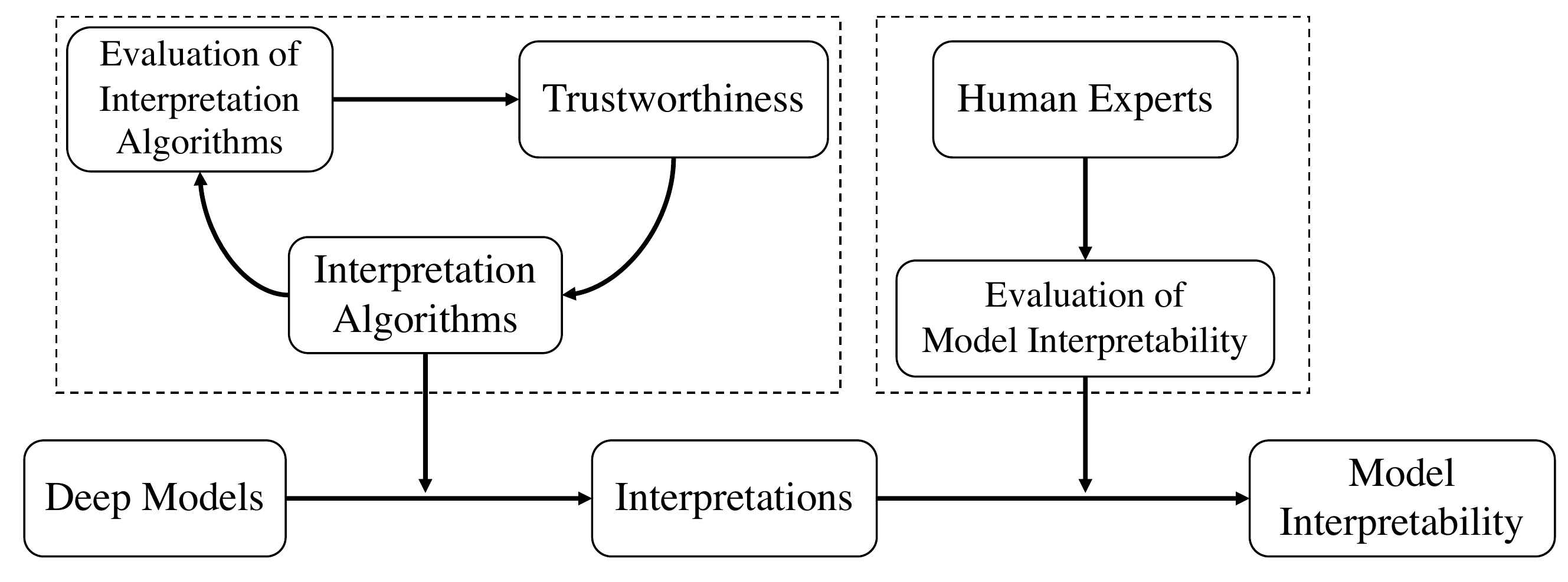}
    \caption{Scheme about interpretations, interpretation algorithms, trustworthiness, model interpretability and the corresponding evaluations.}
    \label{fig:pipeline}
\end{figure}

\paragraph{Interpretation {vs.} Interpretability - } In this work, we first clarify two concepts that should be distinguished: \textit{interpretations} and \textit{model interpretability}.
Interpretations are also named as explanations or attributions that are calculated by \textit{interpretation algorithms} to explain or reveal the ways that deep models make decisions, such as the indication of discriminative features used for model decisions~\cite{DBLP:conf/kdd/Ribeiro0G16}, or the importance of every training sample as the contribution for inference~\cite{DBLP:conf/icml/KohL17}. 
On the other hand, the model interpretability refers to the intrinsic properties of a deep model measuring \emph{in which degree the inference result of the deep model is predictable or understandable to human beings}~\cite{doshi2017towards}.
%
In practice, one could apply the interpretation algorithms of \textit{trustworthiness} (introduced below) to further evaluate the model interpretability through matching the interpretations, i.e., the results from interpretation algorithms for a deep model, with the human-labeled results if available, such as~\cite{DBLP:conf/cvpr/BauZKO017}. 
In this way, the comparison of interpretability becomes possible among different models.
More evaluation approaches are reviewed and will be introduced later.

\paragraph{Interpretation Algorithms and the Taxonomy} \lxh{As there are no formal nor well-agreed definitions about the way to interpret a deep model,} the interpretation algorithms are usually designed with different principles, such as
\begin{itemize}
    \item To highlight the important parts of input features on which the deep model mainly relies, using gradients~\cite{DBLP:journals/corr/SmilkovTKVW17}, perturbations~\cite{DBLP:conf/iccv/FongV17}, proxy explainable models~\cite{DBLP:conf/kdd/Ribeiro0G16} and other methods;
    \item To investigate the inside of deep models to understand the rationale of how models make decisions by visualizing the intermediate features~\cite{DBLP:conf/cvpr/ZhouKLOT16,DBLP:conf/cvpr/ZhangWZ18a}, or putting the counterfactual examples to investigate the changes~\cite{DBLP:conf/icml/GoyalWEBPL19};
    \item To analyze the training data by assessing their individual contributions~\cite{DBLP:conf/icml/KohL17}, estimating their learning difficulty~\cite{DBLP:conf/nips/BaldockMN21} or detecting mislabeled samples~\cite{DBLP:conf/nips/Pleiss0EW20}. 
\end{itemize}
This paper reviews the recent interpretation algorithms and proposes a novel taxonomy for categorizing the interpretation algorithms.
In brief, the proposed taxonomy has three orthogonal dimensions -- (1) \textit{representations of interpretations}, e.g., the input feature importance or the training samples' influences; (2) \textit{the type of the targeting model that the algorithm can be used for}, e.g., differentiable models, models containing specific architectures or other properties; and (3) \textit{relations between interpretation algorithms and the deep model}, e.g., the closed-form expression or the composition of the model.
Recent interpretation algorithms can all be categorized to the proposed three-dimensional taxonomy, which will be presented in detail in Section~\ref{sec:categorization}.



\paragraph{Evaluations on Trustworthiness of Interpretation Algorithms and Model Interpretability} 
There are two evaluations: one on the trustworthiness of interpretation algorithms, another on the model interpretability.

From previous reviews and outlooks for the interpretations~\cite{DBLP:journals/cacm/Lipton18,doshi2017towards,carvalho2019machine,DBLP:conf/acl/JacoviG20,DBLP:journals/pieee/SamekMLAM21}, we summarize the most important desiderata for the interpretation algorithms, i.e., the \textbf{trustworthiness}.
The ``trustworthiness'' here refers to that the interpretation results are reliable/faithful to arbitrary deep models.
That is to say: The trustworthy interpretation algorithm produces the explanations that loyally reveal the model's behaviors, instead of giving results that are irrelevant or just those desired by humans.
Incorporating a trustworthy interpretation algorithm, the evaluations on the model interpretability are then meaningful. 
In Fig.~\ref{fig:pipeline}, we illustrate the connections between these key concepts and further elaborate these concepts in Section~\ref{sec:definition}.

The trustworthiness of the interpretation algorithms could be assessed by designed evaluation approaches for assuring the uses of interpretations, and the interpretability of deep models could be evaluated and measured for identifying the most interpretable ones. 
Both evaluations have challenges remaining, introduced below.
\begin{itemize}
    \item Quantifying the utility of trustworthiness of interpretation algorithms is challenging due to the lack of a proper definition of this quantity and well-defined metrics. Though trustworthiness can be understood subjectively that the trustworthy algorithm produces loyal interpretations to the model, the optimal metric is still under study. Simple metrics such as accuracy, precision, and recall, are not applicable here.
    \item The difficulty of evaluating the model interpretability mainly comes from the lack of the \textit{ground truth}. We could not casually annotate ``true'' interpretations as annotating image labels because interpretation labels might not exist in most cases, or it would be out of objectiveness. Furthermore, obtaining human labeled ground truth for interpretation is labor/time-consuming, which is not scalable over large datasets.
\end{itemize}
Even in this complex and difficult situation, several efficient and effective approaches have been proposed to evaluate the trustworthiness of interpretation algorithms and model interpretability.
The former is mainly based on perturbation evaluations~\cite{DBLP:journals/tnn/SamekBMLM17,DBLP:conf/iclr/HendrycksD19,DBLP:conf/bmvc/PetsiukDS18} or proxy models~\cite{DBLP:conf/iclr/AnconaCO018,DBLP:conf/nips/YehHSIR19}, while the latter based on expert ground truths~\cite{DBLP:conf/cvpr/BauZKO017} or cross-model explanations~\cite{DBLP:journals/corr/abs-2109-00707}.
In Section~\ref{sec:evaluation}, we comprehensively review the evaluation approaches on both the trustworthiness of interpretation algorithms and model interpretability.

\paragraph{Overview}
\lxh{
We describe the organization of this survey paper:
We introduce the key concepts, including the interpretation algorithm, interpretations, model interpretability, and their relations in Section~\ref{sec:definition}.
We present the proposed taxonomy for interpretation algorithms and introduce the algorithms accordingly in Section~\ref{sec:categorization}.
Evaluations on the trustworthiness of interpretation algorithms and the model interpretability are introduced in Section~\ref{sec:evaluation}.
Section~\ref{sec:relations} discusses the connections between interpretations and other research topics.
Finally, we introduce several open-source libraries for interpretations and related in Section~\ref{sec:library}.
}

\section{Main Concepts: Interpretations and Interpretability}
\label{sec:definition}

The fuzziness of main concepts \textit{interpretation} and \textit{interpretability} leads to a lot of confusions and hinders the academic process.
In this section, we make our efforts to clarify these fuzzy research targets and introduce the definitions of \textit{interpretations}, \textit{interpretation algorithms} and \textit{model interpretability}, with involving the notion of \textit{trustworthiness}.

\subsection{Interpretation Algorithms and Trustworthiness}

We first introduce interpretation algorithms.
A deep model needs interpretations because the inference output of the model does not show the reasoning inside.
An interpretation algorithm is thus designed to produce interpretations to explain the model's decisions and gain insight into its internals of reasoning and rationale.
As mentioned previously, there are no formal nor well-agreed definitions about the way to interpret a deep model.
We, therefore, adopt a very loose definition about the interpretation: \textit{All the outcomes produced by the interpretation algorithms that help to understand the model are considered as interpretations}.

\lxh{
Instead of directly discussing the interpretations, we introduce the categories of the interpretation algorithms, as they give different information to help humans to understand the deep models.
For example, an algorithm obtaining the training samples' learning difficulties helps to inspect the model's training process;
An algorithm computing the feature importance helps to realize the most important features that the model uses to make decisions;
An algorithm investigating the intermediate results of a neural network helps understand the model's decision-making process.
We show a novel taxonomy to fully categorize the existing and potential algorithms and review the corresponding algorithms in Section~\ref{sec:categorization}.
}

\lxh{The interpretation can then lead to the discussion that the model is interpretable or not.}
However, before that discussion, we should guarantee at the first step that \lxh{the interpretation algorithm is \textbf{trustworthy} and the interpretation can be trusted.}
The notion of \textbf{trustworthiness} is proposed to cover the most important desiderata from the previous review works~\cite{DBLP:journals/cacm/Lipton18,DBLP:journals/corr/abs-1901-04592,carvalho2019machine}, and can be defined as follows:
\begin{itemize}
    \item \textit{An interpretation algorithm is {trustworthy} if it properly reveals the {underlying rationale} of a model making decisions.}
\end{itemize}
In this definition, the \textit{underlying rationale} \lxh{covers all categories of information that help to understand the model, e.g.}, how the model makes decisions, or the reasoning behind the model making decisions.
The word \textit{properly} here targets the issue that the intrinsic underlying rationale behind the model is usually given by an extrinsic algorithm.
Extrinsic algorithms may not be part of the targeting model to be interpreted.
\lxh{That is to say, as an additional module to diagnose the model, the interpretation algorithm is} at risk of giving explanations that are independent of the the model.
\lxh{A sanity check~\cite{DBLP:conf/nips/AdebayoGMGHK18} was performed to inspect several gradient-based interpretation algorithms by randomizing parts of parameters in the model and showing the interpretation changes.
However, a few algorithms always produce the same interpretations, despite the significant changes of the parameters.}
\lxh{Trustworthiness is defined to recover the rationale of the model, whether the model makes the correct decisions or not, instead of yielding information that is independent of the model.}
\lxh{Though the definition of trustworthiness is not mathematically rigorous, the idea behind is clear.
There are also several evaluations for assessing the trustworthiness, which will be introduced in Section~\ref{subsec:algorithm-evaluation}.
}


\paragraph{Trustworthiness of Different Interpretation Algorithms}
Due to the differences in representation of explanation results and type of models to be interpreted, the amount of information exposed by interpretation algorithms may be different.
Trustworthiness is only required for the explained information.
It would be easy for achieving the trustworthiness if one algorithm explains only a bit of information about the deep model, but this would be rarely useful for any explanation.
The trustworthiness is thus an \textit{ad hoc} requirement with respect to the interpretation algorithm, and defined to guarantee the information provided by the interpretation algorithm can be trusted.


\paragraph{Relation to Self-Interpretable Models}
To complete the discussions of trustworthy interpretation algorithms, we note that many researchers are working on effective \textit{self-interpretable models}, to name a few, Capsule Models~\cite{DBLP:conf/nips/SabourFH17,DBLP:conf/iclr/HintonSF18}, Neural Additive Models~\cite{DBLP:conf/nips/AgarwalMFZLCH21} and CALM~\cite{DBLP:conf/iccv/KimCAO21}.
We consider this is a particular case within our discussion that the self-interpretable models contain both the model and the intrinsic interpretation algorithm.
To be more accurate, the self-interpretable models consist of an intrinsic interpretation algorithm.
Moreover, if the model makes decisions based on the intrinsic interpretations, then this interpretation component is without doubt trustworthy.

\paragraph{Fully-Interpretable Models}
We also discuss \textit{fully interpretable models} here to get a better understanding about the interpretations and the trustworthiness of interpretation algorithms for black-box deep models.
We informally give the definition that a model is fully interpretable if the model is totally understandable by humans.
The following models are considered as fully interpretable without too much controversy\footnote{Without any limits, even a rule-based model may be too complex for a human to understand the model~\cite{DBLP:journals/cacm/Lipton18,rudin2019stop}.
This is also the motivation of several works that pursue the sparsity of explanation results~\cite{DBLP:conf/kdd/Ribeiro0G16}.}: a set of limited number of rules; a depth-limited decision tree; a sparse linear model.

\paragraph{Comparison to Fully and Self Interpretable Models}
To compare across fully interpretable models, self-interpretable models and black-box deep models, we can see: (1) fully interpretable models can be totally understood by showing themselves. 
(2) Self interpretable ones can provide explanations with an amount of information by an intrinsic interpretation algorithm. 
The interpretation algorithms for both fully and self interpretation models are trustworthy. 
(3) For black-box deep models, it is hard to provide such interpretations and much harder to guarantee the the interpretation algorithms be trustworthy.
Fortunately, the interpretations may be different and do not provide the fully interpretable explanation results.
The trustworthiness only guarantees that the amount of information provided by the interpretation algorithm is correct.

\subsection{Model Interpretability}

From industrial demands, the model interpretability is sometimes more important than other metrics such as accuracy because of safety and social issues in domains of autonomous driving, healthcare, criminal justice, financial services and many others.
Though no mathematical definition has been proposed, general agreement about the expression proposed by~\cite{doshi2017towards} has been reached.
We reclaim their definition of model interpretability as follows.
\begin{itemize}
    \item \textit{The model interpretability is the ability (of the model) to explain or to present in understandable terms to a human.}
\end{itemize}
According to other review works~\cite{DBLP:journals/ai/Miller19,carvalho2019machine}, ``\textit{the interpretability of a model is higher if it is easier for a person to reason and trace back why a prediction was made by the model. 
Comparatively, a model is more interpretable than another model if the prior’s decisions are easier to understand than the decisions of the latter.}''


\lxh{
From the definition of the model interpretability, the expression \textit{understandable to a human} is a subjective notion.
It is human-centered~\cite{doshi2017towards,DBLP:journals/corr/abs-1902-00006}, making it complicated to target this research problem of quantitatively measuring and comparing the interpretability of various models.
Till recently, there are not many metrics for quantifying the model interpretability, and Section~\ref{subsec:model-interpretability-evaluation} will introduce the existing evaluation approaches on the model interpretability.
}


We give an intuitive example to show that different models may have different interpretability.
Take image classification~\cite{DBLP:conf/cvpr/DengDSLL009,welinderetal2010caltech} as the task, and a trustworthy algorithm of analyzing the input-output relations as the interpretation algorithm.
We consider two models, and the produced interpretations locate different image pixels.
It is easier to understand if the interpretation aligns with the object parts in the image, while it is harder to understand if the interpretation locates at the background or another accompanied object in the image for recognizing the target object.
Although the trustworthy algorithm reveals the rationales of both models, we prefer the former model because its way of making decisions is more direct to human understandings.


\subsection{Towards Interpretable Deep Learning}

This section defined the trustworthiness of interpretation algorithms and the model interpretability.
We emphasize several points that usually confuse the field with more explicit remarks.

\paragraph{Interpretation Algorithms, Interpretations and Model Interpretability}
The notions of interpretation algorithms, interpretations, and model interpretability should be distinguished. 
Only the interpretability among all these expressions is a property of the model. 
Interpretation algorithms are designed to analyze the black-box model.
Algorithms must be trustworthy; otherwise, the interpretations do not reveal the model's internals.
Their relations and differences are illustrated in Fig.~\ref{fig:pipeline}.

\paragraph{Summary of Desiderata for Interpretations}
In this section, the proposed desiderata is the trustworthiness for interpretation algorithms. Researchers~\cite{DBLP:journals/cacm/Lipton18,doshi2017towards,carvalho2019machine,DBLP:conf/acl/JacoviG20,DBLP:conf/nips/YehHSIR19,DBLP:journals/corr/abs-2110-01167} also proposed many other desiderata for interpretations, interpretation algorithms or interpretability, such as fairness, privacy, reliability, robustness, causality, trust, fidelity, faithfulness, transferability, informativeness, transparency, plausibility, satisfaction, accountability, etc. However, we note that (1) properties, such as informativeness, plausibility, satisfaction, refer to whether the interpretation is understandable to humans, and are different from the trustworthiness in this paper that refers to algorithms; (2) properties, such as reliability, robustness, trust, fidelity, faithfulness, transparency, are similar to trustworthiness or can be comprised by the general definition of trustworthiness; (3) properties, such as causality, transparency, depend on the \textit{underlying rationale} in our context; (4) properties, such as fairness, transferability, privacy, are the standards to constrain the models; and (5) others (e.g., accountability and traceability) are more related to holistic evaluations of the systems. There is slight difference and specific requirements in various scenarios, but the proposed trustworthiness is only for interpretation algorithms. 

\paragraph{Deep Models for High-Dimensional Data for Scientific Discovery}
Though the motivation of interpretations and interpretability at the beginning is to help humans understand the deep models, the interpretations sometimes lead to other valuable and promising findings.
Deep models may be more efficient than humans to cope with high-dimensional data.
From molecules~\cite{DBLP:series/lncs/PreuerKRHU19,jumper2021highly} to black holes~\cite{DBLP:journals/corr/abs-2110-06968}, from chemistry~\cite{DBLP:journals/jcc/GohHV17} to games~\cite{DBLP:journals/nature/SilverSSAHGHBLB17}, deep models could be used to solve many problems.
However, without interpretations, the knowledge discovered by deep models is still unknown for humans, or the scores obtained are not semantic and not fully understood by humans.
Interpretations in these cases could be helpful to find new intelligent patterns and discover new scientific theories.
For example, from a perspective of rationale processes, interpretations can help humans to understand how a model infers;
Or a feature analysis algorithm can help to identify the most important features that the model uses;
Or a tool of investigating the data can help find the typical data samples or the most influential ones that explain how the model makes decisions.
These algorithms are all included in this paper and will be discussed in the following section.

\section{Interpretation Algorithms: Taxonomy, Algorithm Designs, and Miscellaneous}
\label{sec:categorization}

This section introduces the interpretation algorithms in recent years, with a proposed taxonomy of three dimensions.
For each algorithm, we give a brief introduction and follow the taxonomy for the categorization.
A discussion is also provided for future works at the end of this section.

\subsection{Taxonomy}
We categorize the existing interpretation algorithms according to three orthogonal dimensions: representations of interpretations, targeting model's types for interpretations, and the relation between interpretation algorithms and models.
We list the options in each dimension for a better comparison.

For different applications and interpretation requirements, the representations of interpretation are various:
\begin{itemize}[leftmargin=.4in]
    \item \textbf{Feature (Importance)}. These algorithms aim at estimating the feature importance/contribution with respect to the final objective. This includes the analyses on the dimensions of input raw data and extracted features, e.g., images, texts, audios etc.; and intermediate features inside models, e.g., the activations of neural networks; or latent features in GANs. 
    \item \textbf{Model Response}. Algorithms here generally propose to generate or find new examples and see the model's responses, so as to investigate the model behaviors on certain patterns, prototypes, or discriminative features by which the model makes decisions. 
    \item \textbf{Model Rationale Process}. Though deep models are complex, they can be substituted by interpretable models, to gain insights on the rational process inside. Algorithms here interpret the deep model by indicating the path that the model makes decisions.
    \item \textbf{Dataset}. Instead of interpreting deep models, algorithms here propose to explain the data samples in the training set by showing how they affect the optimization phase of deep models. 
\end{itemize}

Interpretation algorithms cope with different types of models:
\begin{itemize}[leftmargin=.4in]
    \item \textbf{Model-agnostic}. Algorithms are included here that completely consider the models as black boxes and do not investigate the inside of models.
    \item \textbf{Differentiable model}. This subset of algorithms contains only algorithms that address the interpretations of differentiable models, especially neural networks. Note that model-agnostic algorithms also cover this subset.
    \item \textbf{Specific model}. This family of algorithms can only be applied to certain types of models, e.g., convolutional neural networks (CNNs), generative adversarial networks (GANs), Graph Neural Networks (GNNs). This is a narrower family than the previous one. 
\end{itemize}

\begin{figure}
    \centering
    \includegraphics[width=\linewidth]{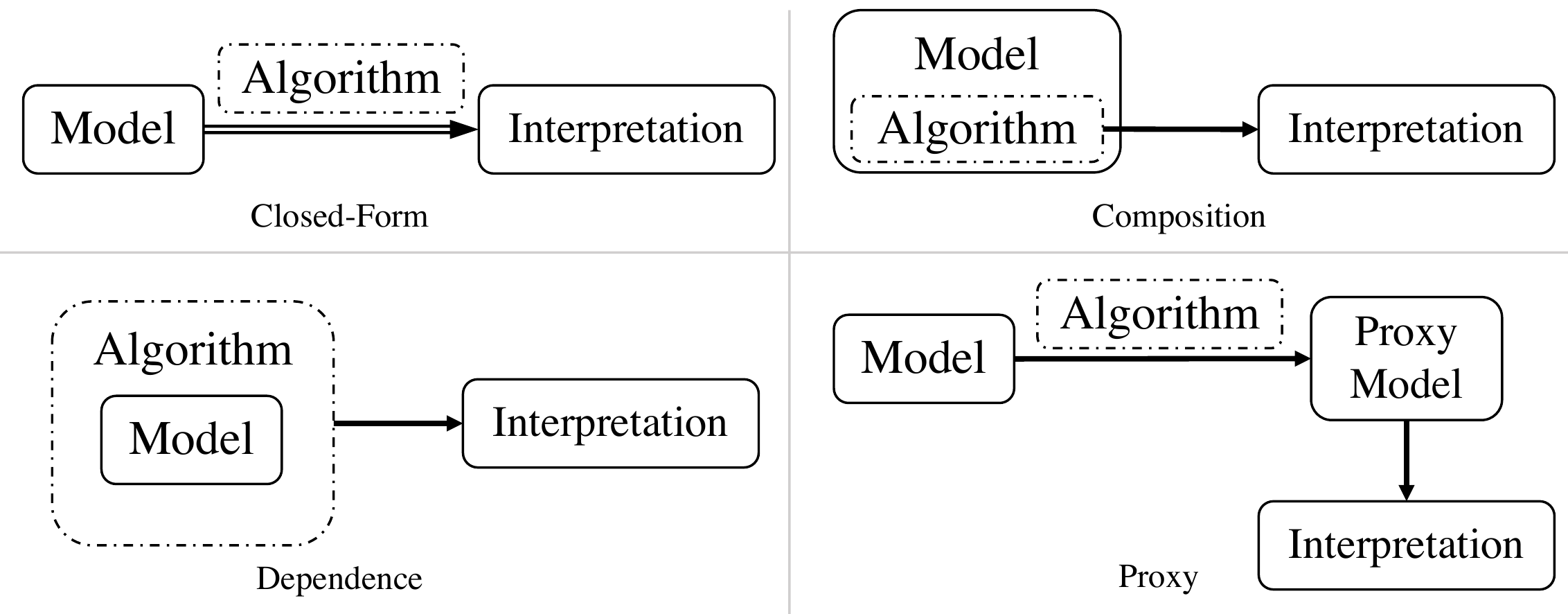}
    \caption{Illustration of relations between the interpretation algorithm and the model. Four relations are illustrated: Closed-form, composition, dependence and proxy.}
    \label{fig:relation-algorithm-model}
\end{figure}
The third dimension for categorizing interpretation algorithms is the relation between the interpretation algorithm and the model:
\begin{itemize}[leftmargin=.4in]
    \item \textbf{Closed-form}. These algorithms derive a closed-form formula from the target model and output interpretable terms.
    \item \textbf{Composition}: Algorithms here can be considered as components of (interpretable) models, usually obtained during training.
    \item \textbf{Dependence}: These algorithms build new operations upon the target model after training and output interpretable terms.
    \item \textbf{Proxy}. Unlike dependence, algorithms here obtain, via learning or derivation, a proxy model for explaining the behavior of models.
\end{itemize}
For a better illustration, four of relations between interpretation algorithms and deep models are shown in Fig.~\ref{fig:relation-algorithm-model}.

We have introduced the proposed taxonomy of three dimensions: Representation, Model Type and the Relation.
In the following subsection, we will present most of the recent interpretation algorithms.
We also give a categorization of all these algorithms with respect to the proposed taxonomy in Table~\ref{table:categorization}.

\subsection{Interpretation Algorithms}
\label{subsec:algos}

\paragraph{LIME and Model-Agnostic Algorithms}
LIME~\cite{DBLP:conf/kdd/Ribeiro0G16} presents a locally faithful explanation by fitting a set of perturbed samples near the target sample using a potentially interpretable model, such as linear models and decision trees. We define a model $g\in G$, where $G$ is a class of interpretable models. The domain of $g$ is $\{0,1\}^{d'}$ and its complexity measure is $\Omega(g)$. Let  $f:\R^d\to	\R$ be the model being explained and $\pi_x(z)$ be the proximity measure between a perturbed sample $z$ and $x$. Finally, let $L(f,g,\pi_x)$ be a measure of the unfaithfulness of $g$ in approximating $f$ in the locality defined by $\pi_x$. LIME produces explanations by the following: 
\begin{equation}
    \xi(x)= \argmin_{g\in G} L(f,g,\pi_x)+\Omega(g).
\end{equation}
The obtained explanation $\xi(x)$ interprets the target sample $x$, with linear weights when $g$ is a linear model.
LIME is model-agnostic, meaning that the obtained proxy model is suitable for any model. 
Similarly, several model-agnostic algorithms, such as Anchors~\cite{DBLP:conf/aaai/Ribeiro0G18}, SHAP~\cite{DBLP:conf/nips/LundbergL17}, RISE~\cite{DBLP:conf/bmvc/PetsiukDS18}, MAPLE~\cite{DBLP:conf/nips/PlumbMT18}, target interpreting features and provide feature importance or contributions to the final decision.

\paragraph{Global Interpretation Algorithms}
Feature importance analysis is a common tool for explaining the model outputs with respect to inputs.
The aforementioned approaches can be categorized into feature importance analysis, while their interpretations are for individual examples, giving unique results for each different example.
Different from these ``local'' interpretations, ``global'' interpretations provide feature importance in an overall vision of the model.
Global interpretations for deep models are usually based on local ones, and an aggregation of local interpretations is performed to obtain the global feature importance, while the difference resides in the aggregation approach, e.g. LIME-SP~\cite{DBLP:conf/kdd/Ribeiro0G16}, NormLIME~\cite{DBLP:journals/corr/abs-1909-04200} and GALE~\cite{DBLP:journals/corr/abs-1907-03039}.

\paragraph{Input Gradients Based Algorithms} The input gradient attributes the important features in the input domain. 
\lxh{However, for deep non-linear models with numerous layers stacked, the gradients would be vanished or saturated during the back-propagation and thus contain noises.}

SmoothGrad~\cite{DBLP:journals/corr/SmilkovTKVW17} proposed to remove the noises \lxh{by averaging the gradients of a number of noised inputs.}
We take visual tasks as an example:
Given input image $x$, neural networks compute a class activation function $S_c$ for class $c\in C$. A sensitivity map can be constructed by calculating the gradient of $M_c$ with respect to input $x$: $M_c(x) = \partial S_c(x)/\partial x$. However, the saliency maps are often noisy because of sharp fluctuations of the derivative. To smooth the gradients, multiple Gaussian noises are added to the input image, and the saliency maps are averaged. SmoothGrad is defined as follows:
\begin{equation}
    \hat{M_c}(x) = \frac{1}{n}\sum_{1}^{n}M_c(x+\mathcal{N}(0,\,\sigma^{2})).
\end{equation}

Integrated Gradient (IG)~\cite{DBLP:conf/icml/SundararajanTY17} aggregates the gradients along with the inputs that lie on the straight line between the baseline and input. Let $F$ be a neural network, $x$ be the input, and $x'$ be the baseline input, which can be a black image for computer vision models and a vector of zeros for word embedding in text models. The integrated gradients along the $i^{th}$ dimension is
\begin{equation}
    \text{IG}_i(x) = (x_i - x_i')\times \int_{\alpha=0}^{1}\frac{\partial F(x'+\alpha \times(x-x'))}{\partial x_i}d\alpha.
\end{equation}
An axiom called \emph{completeness} is satisfied, which states that the attributions add up to the difference between the output of $F$ at input $x$ and baseline $x'$.

Other input gradients based algorithms include DeepLIFT~\cite{DBLP:conf/icml/ShrikumarGK17}, VarGrad~\cite{DBLP:conf/nips/AdebayoGMGHK18}, GradSHAP~\cite{DBLP:conf/nips/LundbergL17}, and FullGrad~\cite{DBLP:conf/nips/SrinivasF19}.

\paragraph{Layer-wise Relevance Propagation}
Layer-wise relevance propagation (LRP)~\cite{bach2015pixel} is also an input feature attribution algorithm.
Instead of using proxy models, perturbations or gradients, LRP recursively computes a Relevance score for each neuron of layers, so as to understand the contribution of a single pixel of an image $x$ to the prediction function $f(x)$ in an image classification task.
\begin{align}
    \label{eq:lrp}
    f(x) = \cdots = \sum_{d=1}^{V^{(l+1)}} R_d^{(l+1)} = \sum_{d=1}^{V^{(l)}} R_d^{(l)} = \cdots = \sum_{d=1}^{V^{(1)}} R_d^{(1)},
\end{align}
where $R_d^{(l)}$ is the Relevance score of the $d$-th neuron at the $l$-th layer, $V^{(l)}$ indicates the dimension of $l$-th layer, and $V^{(1)}$ is the number of pixels in the input image.
Iterating Eq. (\ref{eq:lrp}) from the last layer, which is the classifier output $f(x)$ to the input layer $x$ consisting of image pixels, then yields the contribution of pixels to the prediction results.
Based on the idea of back-propagating Relevance scores, LRP can be extended to other neural networks, even with special and complex nonlinear operations~\cite{DBLP:conf/icann/BinderMLMS16,DBLP:journals/pr/MontavonLBSM17}.
To adapt LRP to specific tasks, many variants have been proposed, such as Contrastive LRP~\cite{DBLP:conf/accv/GuYT18} which produces pixel-wise explanations of instance objects, Softmax-Gradient LRP~\cite{DBLP:conf/iccvw/IwanaKU19} which gives explanations focusing on discriminating possible objects in the images, and Relative Attributing Propagation (RAP)~\cite{DBLP:conf/aaai/NamGCWL20} which focuses on both positive and negative features.
Furthermore, extended LRPs~\cite{DBLP:conf/acl/VoitaTMST19,DBLP:conf/cvpr/CheferGW21} can be helpful to interpret Transformer models~\cite{DBLP:conf/naacl/DevlinCLT19,DBLP:journals/corr/abs-1904-09223,DBLP:conf/iclr/DosovitskiyB0WZ21}.


\paragraph{CAM and Variants} 
Given a CNN and an image classification task, classification activation map (CAM)~\cite{DBLP:conf/cvpr/ZhouKLOT16} can be derived from the operations at the last layers of the CNN model and show the important regions that affect model decisions.
Specifically, for a given category $c$, we expect the unit corresponding to a pattern of the category in the receptive field be activated in the feature map. The weights in the classifier indicate the importance of each feature map in classifying category $c$. Therefore, a weighted sum of visual patterns illustrates the important regions of a category. Let $f_k(x,y)$ denote the activation of unit k in the last convolutional layer at spatial location $(x,y)$, $F_k=\sum_{x,y}f_k(x,y)$ be the global average pooling for unit k, and $w_k^c$ be the weight corresponding to class $c$ for unit $k$ so that $\sum_{k}w_k^c F_k$ is the input to softmax for class $c$. Then the activation map for class c is:
\begin{equation}
    M_c(x,y) = \sum_{k}w_k^c f_k(x,y).
\end{equation}

GradCAM~\cite{DBLP:journals/ijcv/SelvarajuCDVPB20} further looks at the gradients flowing into the convolutional layer to give weight to activation maps. Let $y^c$ be the score for class c before the softmax, $A^k$ be feature map activations of the unit k in a convolutional layer, the neuron importance weight $\alpha_{k}^c$ is the global-average-pooled gradient of $y^c$ with respect to $A^k$:
\begin{equation}
    \alpha_{k}^c = \frac{1}{Z}\sum_{i}\sum_{j}\frac{\partial y^c}{\partial A_{i,j}^k}.
\end{equation}
The localization map is a weighted combination of activation maps:
\begin{equation}
    L_{Grad-CAM}^c = ReLU(\sum_{k}\alpha_k^c A^k).
\end{equation}

ScoreCAM~\cite{DBLP:conf/cvpr/WangWDYZDMH20} also uses gradient information but assigns importance to each activation map by the notion of \emph{Increase of Confidence}. Given an image model $Y=f(X)$ that takes in image X and outputs logits Y. The $k$-th channel of convolutional layer $l$ is denoted $A_l^k$. With baseline image $X_b$ and category c, the contribution $A_l^k$ towards Y is:
\begin{equation}
    C(A_l^k) = f^c(X\circ H_l^k) - f^c(X_b),
\end{equation}
where $H_l^k = s(Up(A_l^k))$. $Up(\cdot)$ is the operation that upsamples $A_l^k$ into the input size and $s$ normalizes each element into $[0,1]$.
ScoreCAM is defined as:
\begin{equation}
    L_{Score-CAM}^c = ReLU(\sum_{k}\alpha_k^c A_l^k), 
\end{equation}
where $\alpha_k^c = C(A_l^k)$.

More CAM variants have been recently proposed, e.g., GradCAM++~\cite{DBLP:conf/wacv/ChattopadhyaySH18}, CBAM~\cite{DBLP:conf/eccv/WooPLK18}, Respond-CAM~\cite{DBLP:conf/miccai/ZhaoZWJX18}, and Ablation-CAM~\cite{DBLP:conf/wacv/DesaiR20}.


\paragraph{Perturbation-Based Algorithms} To investigate important features in the input, a straightforward way is to measure the effect of perturbations applied to the input~\cite{DBLP:conf/iccv/FongV17,DBLP:conf/iccv/FongPV19}.
This idea is quite simple: The random perturbations on the features would lead to different changes in the model's predictions, where larger changes would be observed for more important features.
Note that perturbation can be also used for evaluating the trustworthiness of interpretation algorithms when we are not aware of interpretation ground truth~\cite{DBLP:journals/tnn/SamekBMLM17,DBLP:journals/corr/abs-1906-02032}.

\paragraph{Counterfactual Examples}
Using counterfactual examples to explain the model behaviors is also an important direction for understanding the black boxes.
Generally, the counterfactual examples have changes in the input that are as small as possible, but would completely change the decision made by the model.
The changes in input would be a clue for explaining the model's behavior.
Most counterfactual-example approaches, such as FIDO~\cite{DBLP:conf/iclr/ChangCGD19}, DiCE~\cite{DBLP:conf/fat/MothilalST20}, and several others~\cite{DBLP:conf/pkdd/LaugelLMRD19,DBLP:conf/icml/GoyalWEBPL19}, to generating counterfactual examples are based on the optimization with sparsity constraints or towards the smallest changes in input.
Using counterfactual examples to explain the model behaviors can also be included in causal inference~\cite{pearl2009causal}, which is considered as a new perspective for model interpretability \cite{DBLP:journals/sigkdd/MoraffahKGRL20,DBLP:journals/corr/abs-2006-16789}.
Detailed reviews on counterfactual explanations can be found in~\cite{DBLP:journals/corr/abs-1711-00399,DBLP:journals/corr/abs-1911-07749,DBLP:journals/corr/abs-2010-10596}. 

\paragraph{Adversarial Examples}
Adversarial examples are very related to counterfactual ones with similar optimization methods, while adversarial examples are used to reveal the vulnerability of the deep model and often attack the AI systems.
Adversarial examples in vision tasks, are usually the imperceptible changes in the images which mislead the model's decision.
Note that analyses on the adversarial examples~\cite{DBLP:conf/nips/IlyasSTETM19,DBLP:journals/natmi/GeirhosJMZBBW20} show the connections to the understanding of the deep learning process and robustness of the trained deep model.

\paragraph{TCAV}
Given a set of examples representing a concept of human interest (such as an object, a pattern, a color etc.), TCAV~\cite{DBLP:conf/icml/KimWGCWVS18} seeks a vector in the space of activations at some layer to represent this concept.
Precisely, by defining a concept activation vector (or CAV) as the normal to a hyperplane, TCVA separates examples according to the existence of this concept in the activations:
Given one example in a particular class, along the direction of a CAV, the directional derivative of this example contributes a score if it is positive, and the ratio of examples that have positive directional derivatives over all examples in this class is defined as the TCAV score.
CAV finds examples of a semantic concept learned by the intermediate layers of a deep model, contributing to the predictions while TCAV quantitatively measures the contributions of this concept.

\paragraph{Prototype}
To explain the classification models, finding the typical exemplar for each category is also effective and direct.
Humans can understand better that the model identifies the featured prototype to make decisions.
Chen et al.~\cite{DBLP:conf/nips/ChenLTBRS19} proposed ProtoPNet, which explains the deep model by finding prototypical parts of predicted objects and gathering evidence from the prototypes to make final decisions.
Another method named ABELE~\cite{DBLP:conf/pkdd/GuidottiMMP19} generates exemplar and counter-exemplar images, labeled with the class identical to, and different from, the class of the image to explain, with a saliency map, highlighting the importance of the areas of the image contributing to its classification.

As a technique for generating prototypes, activation maximization generally computes the prototypes through an optimization process:
\begin{align}
    \max_{\vx} \ \log p(y_c|\vx) - \lambda \| \vx \|^2,
\end{align}
where $p(y_c|\vx)$ is the probability given by a deep model with $\vx$ as input, and the second term is the constraint for generating the prototype.
However, the constraint can be replaced by many other choices~\cite{erhan2009visualizing,DBLP:journals/corr/SimonyanVZ13,DBLP:conf/cvpr/MahendranV15,DBLP:conf/nips/NguyenDYBC16}.
A tutorial for this direction is cited~\cite{DBLP:journals/dsp/MontavonSM18}.
More works related to prototypes or exemplars for interpretations can be found in \cite{bien2011prototype,DBLP:conf/aaai/LiLCR18,DBLP:conf/kdd/MingXQR19,DBLP:conf/pkdd/BarbalauCIP20}.

\paragraph{Proxy Models for Rationale Process}
The reasoning process or the underlying rationale of deep models is complex due to the non-linearity and enormous computations.
It is difficult for humans to know the exact steps of the rationale process with semantics inside the black boxes.
However, this rationale process can be proxied by graph models~\cite{DBLP:conf/aaai/ZhangCSWZ18} or decision trees~\cite{DBLP:conf/cvpr/ZhangYMW19}, which provide a decision-making path that is more interpretable to humans.
Moreover, deep neural networks can be combined with decision forest models ~\cite{DBLP:conf/iccv/KontschiederFCB15} or distilled into a soft decision tree~\cite{DBLP:conf/aiia/FrosstH17}.
A model-agnostic approach for interpreting rationale process named BETA~\cite{DBLP:journals/corr/LakkarajuKCL17} allows to learn (with optimality guarantees) a small number of compact decision sets, each of which explains the behavior of the black box model in specific, well-defined regions of feature space.


\paragraph{Forgetting Events}
Forgetting events are defined by \cite{DBLP:conf/iclr/TonevaSCTBG19} for analyzing the training examples using training dynamics.
Given a dataset $D=(x_i,y_i)_i$, after $t$ steps of SGD, example $x_i$ undergoes a forgetting event if it is misclassified at step $t+1$ after having been correctly classified at step $t$. Forgetting events signify samples' interactions with decision boundaries, and the samples play a part equivalent to support vectors in the \emph{support vector machine} paradigm. Unforgettable examples are samples learnt at step $t^*<\infty$ and never misclassified for all $k \geq t^*$. They are easily recognizable samples that contain obvious class attributes. Whereas examples with the most forgetting events are ambiguous without clear characteristics of a certain class, and some are noisy samples. 

\paragraph{Dataset Cartography}
Dataset cartography~\cite{DBLP:conf/emnlp/SwayamdiptaSLWH20} looks into two measures for each sample during the training process - the model's confidence in the true class and the variability of confidence across epochs. Therefore, training examples can be categorized as easy-to-learn, hard-to-learn, or ambiguous based on their position in the two-dimensional map. Consider training dataset $D={(x,y^*)_i}_{i=1}^N$ where $x_i$ is the $i$-th sample and $y_i^*$ is the true label. After training for $E$ epochs, the confidence is defined as the mean probability of true label across epochs:
\begin{equation}
    \hat{\mu_i} = \frac{1}{E}\sum_{e=1}^{E}p_{\theta^{(e)}}(y_i^*|x_i),
\end{equation}
where $p_{\theta^{(e)}}$ is the probability with parameters $\theta^{(e)}$ at the end of the $e^{th}$ epoch.
The variability is the standard deviation of $p_{\theta^{(e)}}(y_i^*|x_i)$:
\begin{equation}
    \hat{\sigma_i} = \sqrt{\frac{\sum_{e=1}^{E}(p_{\theta^{(e)}}(y_i^*|x_i)-\hat{\mu_i})^2}{E}},
\end{equation}

\paragraph{AUM}
Another method for analysing the training dynamics is proposed to compute the area under the margin (AUM)~\cite{DBLP:conf/nips/Pleiss0EW20}:
\begin{align}
    \text{AUM}(\vx, y) = \frac{1}{T} \sum_{t=1}^T ( z_y^{(t)}(\vx) - \max_{i \neq y} z_i^{(t)}(\vx) ),
\end{align}
where $z_i^{(t)}(\vx)$ is the logit, computed by the model, of $i$-th class at $t$-th epoch during training with respect to the example $\vx$.

\paragraph{Influence Functions}
Influence functions~\cite{DBLP:conf/icml/KohL17} identify the training samples most responsible for a model prediction by upweighting a sample by some small value and analyze its effect on the parameters and the loss of the target sample. Given input space $X$ and output space $Y$, we have training data $z_1, \dots, z_n$, where $z_i=(x_i,y_i)\in X\times Y$. Let $L(z, \theta)$ be the loss where $\theta\in\Theta$ are the parameters. The optimal $\hat{\theta}$ is given by $\hat{\theta}=argmin_{\theta\in\Theta}\frac{1}{n}\sum_{i=1}^{n}L(z_i,\theta)$. The influence of upweighting training point $z$ on the loss at the test point $z_{test}$ is: 
\begin{equation}
    I_{up,loss}(z,z_{test})=-\nabla_\theta L(z_{test},\hat{\theta})^T {H_{\hat{\theta}}}^{-1} \nabla_\theta L(z,\hat{\theta}),
\end{equation}
where $H_{\hat{\theta}}=\frac{1}{n}\sum_{i=1}^{n}\nabla_\theta^2L(z_i,\hat{\theta})$.
Based on influence functions, several techniques~\cite{DBLP:conf/nips/KohATL19,DBLP:conf/aaai/ChenLYWM21} have been proposed with improvement.

\paragraph{Contributions of Long-Tailed Training Examples}
Instead of identifying mislabeled samples, easy/difficult-to-learn samples from the training set, more theoretical works on detecting the long-tail examples and outliers~\cite{DBLP:conf/nips/FeldmanZ20,DBLP:journals/corr/abs-1910-13427,DBLP:conf/stoc/Feldman20}.
Most of them investigate the connections between the memorization capacity of deep models~\cite{DBLP:journals/cacm/ZhangBHRV21} and the learning process, in order to know the contributions of training examples, including long-tailed ones and outliers.

\paragraph{Interpretations on GNNs} 
Graph Neural Networks (GNNs) are a powerful tool for learning tasks on structured graph data. 
Like other deep learning models, GNNs show the black-box fashion and are required to explain their prediction results and rationale processes.
Without requiring modification of the underlying GNN architecture, GNNExplainer~\cite{DBLP:conf/nips/YingBYZL19} leverages the recursive neighborhood-aggregation scheme to identify important graph pathways as well as highlight relevant node feature information that is passed along edges of the pathways.
Recently, more researches focus on the interpretations of GNN models, such as GraphLIME~\cite{DBLP:journals/corr/abs-2001-06216}, CoGE~\cite{DBLP:journals/corr/abs-2010-13663}, Counterfactual explanations on GNNs~\cite{DBLP:conf/nips/BajajCXPWLZ21} and others~\cite{DBLP:journals/corr/abs-1905-13686,DBLP:conf/cvpr/PopeKRMH19,DBLP:conf/nips/LuoCXYZC020}.

\paragraph{Interpretations on GANs}
Generative adversarial networks (GANs) are a popular generative model based on two adversarial networks, where one generates synthesized examples, and another tries to classify generated examples from natural examples.
Interpretations on GANs mainly search for semantically meaningful directions.
Compared with labeled semantics, Bau et al.~\cite{DBLP:conf/iclr/BauZSZTFT19} proposed GAN dissection to find semantic neurons in generative models and modify the semantics in the generated images.
Instead of relying on labels, Voynov et al.~\cite{DBLP:conf/icml/VoynovB20} found semantically meaningful directions in an unsupervised way from the intermediate layers of generative models.
Similarly, Shen et al.~\cite{DBLP:conf/cvpr/ShenZ21} proposed a closed-form factorization method for identifying semantic neurons.
Note that there are other methods for explaining the generative models~\cite{DBLP:journals/ijcv/YangSZ21,DBLP:journals/corr/abs-1912-10920,DBLP:conf/iclr/PlumeraultBH20}.

\paragraph{Information Flow} In some deep learning models there are multiplicative scalar weights that control information flow in some parts of a network. The most common examples are attention~\cite{DBLP:journals/corr/BahdanauCB14} and gating:
\begin{align}
    c^{att} = \sum_i \alpha^{att}_i h_i, \qquad c^{gate} = \alpha^{gate} h
\end{align}
The attention weights $\alpha^{att}$ ($\sum_i \alpha^{att}_i = 1$) and the gate values $\alpha^{gate}$ ($\alpha^{gate} \in [0,1]$) are usually interpretable because their values represent the strength of the corresponding information pathways. 
Attention and gating are frequently used in NLP models, and there have been plenty of works aiming to understand the model through these weights, such as Rollout~\cite{DBLP:conf/acl/AbnarZ20}, Seq2Seq-Vis~\cite{DBLP:journals/tvcg/StrobeltGBPPR19} and others~\cite{DBLP:journals/tvcg/StrobeltGPR18, DBLP:conf/emnlp/GhaeiniFT18}, or to investigate the reliability of using them as explanations~\cite{DBLP:conf/acl/SerranoS19, DBLP:conf/naacl/JainW19, DBLP:conf/emnlp/WiegreffeP19}.
As well, these ideas have also been used in Vision Transformers~\cite{DBLP:conf/iclr/DosovitskiyB0WZ21} for explaining image classification models~\cite{DBLP:conf/cvpr/CheferGW21,yuan2021explaining} or bi-modal transformer models~\cite{DBLP:conf/iccv/CheferGW21}.

\paragraph{Self-Generated Explanations} Using text generation techniques, a model can explicitly generate human-readable explanations for its own decision. A joint output-explanation model is trained to produce an prediction and simultaneously generate an explanation for the reason of that prediction~\cite{DBLP:conf/acl/AtanasovaSLA20, DBLP:conf/acl/LiuYW19, DBLP:conf/acl/KumarT20}. This requires some kind of supervision available to train the explanation part of the model.

\paragraph{Inductive Biases Towards Interpretation Modules}
Different from post-hoc explanations after the optimization process, some works focus on designing inductive biases during training to encourage the model to be more interpretable.
By simple abstraction, the objective function for this purpose can be written as
\begin{align}
    Loss = L(f(x), y) + \alpha R,
\end{align}
where $f(x)$ represents the deep model output with $x$ as input, y is the ground truth, $L$ is the loss function, specifically the cross entropy for standard supervised classification problem, and $R$ is the objective function for biasing towards interpretable models.
Various approaches~\cite{DBLP:conf/cvpr/DongSZZ17,DBLP:conf/aaai/RossD18,DBLP:conf/cvpr/ZhangWZ18a,DBLP:journals/corr/abs-2012-01166} have been proposed to improve the interpretability during training.
More encouragingly, Sabour et al.~\cite{DBLP:conf/nips/SabourFH17} designed a self-interpretable deep model where each neuron outputs semantic features.

\begin{table}[t]
\renewcommand{\arraystretch}{1.25}
\caption{Categorization of interpretation algorithms with respect to the proposed taxonomy. Algorithms are listed following the order of presentation in Section~\ref{subsec:algos}. Note that, each row may contain several algorithms which they may target at explaining different types of models or have different relations to the models. Here the publications in each category of algorithms can be found in the corresponding paragraphs, and are not repeated for a compact table presentation.}
\begin{tabular}{|l|r|r|r|}
\hline
\textbf{Algorithms}  & \textbf{Representation} & \textbf{Model Type}                                                          & \textbf{Relation}                                                    \\ \hline
LIME and Variants                                                                                                   & Feature                 & Model-Agnostic                                                               & Proxy                                                                \\ \hline
Global Interpretation                                                                                                     & Feature                 & Model-Agnostic                                                               & Proxy                                                                \\ \hline
Input-Gradient Based                                                                                                      & Feature                 & Differentiable                                                               & Dependence                                                           \\ \hline
LRP and Variants                                                                                                          & Feature                 & Differentiable                                                               & Dependence                                                           \\ \hline
CAM and Variants                                                                                                          & Feature                 & \begin{tabular}[c]{@{}r@{}}Specific (CNNs) \\[-0.25em] or Differentiable\end{tabular} & \begin{tabular}[c]{@{}r@{}}Closed-Form \\[-0.25em] or Dependence\end{tabular} \\ \hline
Perturbation-Based                                                                                                        & Feature                 & Model-Agnostic                                                               & Dependence                                                           \\ \hline
Counterfactual Examples                                                                                                   & Response                & \begin{tabular}[c]{@{}r@{}}Model-Agnostic \\[-0.25em] or Differentiable\end{tabular}  & Dependence                                                           \\ \hline
Adversarial Examples                                                                                                      & Response                & \begin{tabular}[c]{@{}r@{}}Model-Agnostic \\[-0.25em] or Differentiable\end{tabular}  & Dependence                                                           \\ \hline
TACV                                                                                                                      & Feature                 & Differentiable                                                               & Proxy                                                                \\ \hline
Prototype-Based                                                                                                                & Response                & \begin{tabular}[c]{@{}r@{}}Model-Agnostic \\[-0.25em] or Differentiable\end{tabular}  & Proxy                                                                \\ \hline
\begin{tabular}[c]{@{}l@{}}Proxy Models for \\[-0.25em] Rationale Process\end{tabular}                                             & Rationale               & Specific (CNNs)                                                              & Proxy                                                                \\ \hline
Training Dynamics Based & Dataset                 & Model-Agnostic                                                               & Dependence                                                           \\ \hline
\begin{tabular}[c]{@{}l@{}}Influence Functions \\[-0.25em] and Variants\end{tabular}                                                                                                     & Dataset                 & Differentiable                                                               & \begin{tabular}[c]{@{}r@{}}Closed-Form \\[-0.25em] or Dependence\end{tabular} \\ \hline
\begin{tabular}[c]{@{}l@{}}Contributions of \\[-0.25em] Training Examples\end{tabular}                                             & Dataset                 & Differentiable                                                               & Dependence                                                           \\ \hline
Interpretations on GNNs                                                                                                   & Feature                 & Specific (GNNs)                                                              & Dependence                                                           \\ \hline
Interpretations on GANs                                                                                                   & Feature                 & Specific (GANs)                                                              & Dependence                                                           \\ \hline
Information Flow                                                                                                          & Feature                 & \begin{tabular}[c]{@{}r@{}}Specific \\[-0.25em] (Transformers)\end{tabular}           & Dependence                                                           \\ \hline
Self-Generated Explanations                                                                                               & Feature                 & Specific (NLP)                                                               & Composition                                                          \\ \hline
Self-Interpretable Models  & Rationale               & \begin{tabular}[c]{@{}r@{}}Specific \\[-0.25em] (Self-Interpretable)\end{tabular}     & Composition                                                          \\ \hline
\end{tabular}
\label{table:categorization}
\end{table}

\begin{table}[t]
\renewcommand{\arraystretch}{1.25}
\caption{List of interpretation algorithm publications. Algorithms are listed following the order of presentation in Section~\ref{subsec:algos}.}
\begin{tabular}{|l|l|}
\hline
\multicolumn{1}{|l|}{\textbf{Methods}}                                        & \multicolumn{1}{c|}{\textbf{Publications (Non-Exhaustive)}}                 \\ \hline
LIME and Variants                                                             & \begin{tabular}[c]{@{}l@{}}LIME~\cite{DBLP:conf/kdd/Ribeiro0G16}, Anchors~\cite{DBLP:conf/aaai/Ribeiro0G18}, SHAP~\cite{DBLP:conf/nips/LundbergL17}, \\[-0.25em] RISE~\cite{DBLP:conf/bmvc/PetsiukDS18}, MAPLE~\cite{DBLP:conf/nips/PlumbMT18}\end{tabular}   \\ \hline
Global Interpretation                                                         & LIME-SP~\cite{DBLP:conf/kdd/Ribeiro0G16}, NormLIME~\cite{DBLP:journals/corr/abs-1909-04200}, GALE~\cite{DBLP:journals/corr/abs-1907-03039}                                              \\ \hline
Input-Gradient Based                                                          & \begin{tabular}[c]{@{}l@{}}SmoothGrad~\cite{DBLP:journals/corr/SmilkovTKVW17}, IG~\cite{DBLP:conf/icml/SundararajanTY17}, DeepLIFT~\cite{DBLP:conf/icml/ShrikumarGK17}, \\[-0.25em] VarGrad~\cite{DBLP:conf/nips/AdebayoGMGHK18}, GradSHAP~\cite{DBLP:conf/nips/LundbergL17}, FullGrad~\cite{DBLP:conf/nips/SrinivasF19}\end{tabular}           \\ \hline
LRP and Variants                                                              & \begin{tabular}[c]{@{}l@{}}LRP~\cite{bach2015pixel,DBLP:conf/icann/BinderMLMS16,DBLP:journals/pr/MontavonLBSM17}, Contrastive LRP~\cite{DBLP:conf/accv/GuYT18}, \\[-0.25em] Softmax-Gradient LRP~\cite{DBLP:conf/iccvw/IwanaKU19}, RAP~\cite{DBLP:conf/aaai/NamGCWL20}, \\[-0.25em] Chefer et al.~\cite{DBLP:conf/cvpr/CheferGW21}\end{tabular} \\ \hline
CAM and Variants                                                              & \begin{tabular}[c]{@{}l@{}}CAM~\cite{DBLP:conf/cvpr/ZhouKLOT16}, GradCAM~\cite{DBLP:journals/ijcv/SelvarajuCDVPB20}, ScoreCAM~\cite{DBLP:conf/cvpr/WangWDYZDMH20}, \\[-0.25em] GradCAM++~\cite{DBLP:conf/wacv/ChattopadhyaySH18}, CBAM~\cite{DBLP:conf/eccv/WooPLK18}, \\[-0.25em] Respond-CAM~\cite{DBLP:conf/miccai/ZhaoZWJX18}, Ablation-CAM~\cite{DBLP:conf/wacv/DesaiR20} \end{tabular} \\ \hline
Perturbation-Based                                                            & Fong et al.~\cite{DBLP:conf/iccv/FongPV19,DBLP:conf/iccv/FongV17}, Samek et al.~\cite{DBLP:journals/tnn/SamekBMLM17}, Vu et al.~\cite{DBLP:journals/corr/abs-1906-02032},                                \\ \hline
Counterfactual Examples                                                       & FIDO~\cite{DBLP:conf/iclr/ChangCGD19}, DiCE~\cite{DBLP:conf/fat/MothilalST20}, Goyal et al.~\cite{DBLP:conf/icml/GoyalWEBPL19}, Laugel et al.~\cite{DBLP:conf/pkdd/LaugelLMRD19}                                                           \\ \hline
Adversarial Examples                                                          & Geirhos et al.~\cite{DBLP:journals/natmi/GeirhosJMZBBW20}, Ilyas et al.~\cite{DBLP:conf/nips/IlyasSTETM19}                                          \\ \hline
TACV                                                                          & TACV~\cite{DBLP:conf/icml/KimWGCWVS18}                                                                 \\ \hline
Prototype-Based                                                               & ProtoPNet~\cite{DBLP:conf/nips/ChenLTBRS19}, ABELE~\cite{DBLP:conf/pkdd/GuidottiMMP19}                                                     \\ \hline
\begin{tabular}[c]{@{}l@{}}Proxy Models for \\[-0.25em] Rationale Process\end{tabular} & Zhang et al.~\cite{DBLP:conf/aaai/ZhangCSWZ18,DBLP:conf/cvpr/ZhangYMW19}, BETA~\cite{DBLP:journals/corr/LakkarajuKCL17}                                                   \\ \hline
Training Dynamics Based                                                       & \begin{tabular}[c]{@{}l@{}}Forgetting Events~\cite{DBLP:conf/iclr/TonevaSCTBG19}, Datasets Cartography~\cite{DBLP:conf/emnlp/SwayamdiptaSLWH20}, \\ AUM~\cite{DBLP:conf/nips/Pleiss0EW20}\end{tabular} \\ \hline
\begin{tabular}[c]{@{}l@{}}Influence Functions \\[-0.25em] and Variants\end{tabular}   &  \begin{tabular}[c]{@{}l@{}}Influence Functions~\cite{DBLP:conf/icml/KohL17}, Group Influences~\cite{DBLP:conf/nips/KohATL19},\\[-0.25em] HYDRA~\cite{DBLP:conf/aaai/ChenLYWM21}  \end{tabular}                    \\ \hline
\begin{tabular}[c]{@{}l@{}}Contributions of \\[-0.25em] Training Examples\end{tabular} & Carlini et al.~\cite{DBLP:journals/corr/abs-1910-13427}, Feldman et al.~\cite{DBLP:conf/stoc/Feldman20,DBLP:conf/nips/FeldmanZ20}                                                                     \\ \hline
Interpretations on GNNs                                                       & GNN Explainer~\cite{DBLP:conf/nips/YingBYZL19}, GraphLIME~\cite{DBLP:journals/corr/abs-2001-06216}, CoGE~\cite{DBLP:journals/corr/abs-2010-13663}                                       \\ \hline
Interpretations on GANs                                                       & \begin{tabular}[c]{@{}l@{}}GAN Dissection~\cite{DBLP:conf/iclr/BauZSZTFT19}, Voynov et al.~\cite{DBLP:journals/corr/abs-1912-10920,DBLP:conf/icml/VoynovB20}, \\[-0.25em] Shen et al.~\cite{DBLP:conf/cvpr/ShenZ21}\end{tabular} \\ \hline
Information Flow                                                              & \begin{tabular}[c]{@{}l@{}}Rollout~\cite{DBLP:conf/acl/AbnarZ20}, Seq2Seq-Vis~\cite{DBLP:journals/tvcg/StrobeltGBPPR19}, \\[-0.25em] Chefer et al.~\cite{DBLP:conf/cvpr/CheferGW21,DBLP:conf/iccv/CheferGW21}, TAM~\cite{yuan2021explaining}\end{tabular} \\ \hline
Self-Generated Explanations                                                   & Atanasova et al.~\cite{DBLP:conf/acl/AtanasovaSLA20}, Kumar et al.~\cite{DBLP:conf/acl/KumarT20}, Liu et al.~\cite{DBLP:conf/acl/LiuYW19}                          \\ \hline
Self-Interpretable Models                                                     & Capsule~\cite{DBLP:conf/nips/SabourFH17,DBLP:conf/iclr/HintonSF18}, Neural Additive Models~\cite{DBLP:conf/nips/AgarwalMFZLCH21}, CALM~\cite{DBLP:conf/iccv/KimCAO21}                                \\ \hline
\end{tabular}
\label{table:publication}
\end{table}

\subsection{Categorization and Discussion}

We have introduced a large number of typical interpretation algorithms and categorized them according to the proposed taxonomy, so as to provide a clear illustration in this research field. 
We hope the taxonomy can shed light on future improvements/extensions on explaining (deep) learning models.
We show the categorization of all these algorithms with respect to the proposed taxonomy in Table~\ref{table:categorization}, and gathering interpretation algorithms according to the categorization in Table~\ref{table:publication} for a quick glimpse.

Table~\ref{table:categorization} shows that there are many methods of  the \textbf{Feature} representation and only a few \textbf{Rational} ones; many \textbf{Proxy} and \textbf{Dependence relations} but a few \textbf{Closed-Form}.
We argue that both of these observations were due to the challenging analyses of complex deep neural networks.
The rationale and the closed-form of deep models are still hard to understand or even approximate.
From the categorization, we also would like to point out the blanks that may indicate some unexplored directions for future perspectives.
For example, no \textbf{Model-Agnostic} algorithms have the \textbf{Composition} relation with models.
While the input-output sensitivity analysis methods are currently developed, improving the input-output interpretations can be a good perspective.
Moreover, we should note that the adversarial attacks do not only aim at trained models~\cite{DBLP:journals/corr/abs-1810-00069}, but the interpretations~\cite{DBLP:conf/nips/HeoJM19,DBLP:journals/corr/abs-1902-03501,DBLP:journals/corr/abs-1806-08049}.
We leave the further investigations for future work.

	

\subsection{Interpretations on Specific Application Domains}

We do not explicitly categorize the interpretation algorithms according to their application domains because (1) the algorithm used in one specific domain may also be applicable on a broader scope with little modifications, especially for model-agnostic algorithms; and (2) for model-specific algorithms, the categorization on the model type generally overlaps with the one on the application domain.
For completeness, we discuss recent works of deep model interpretations in the following domains: reinforcement learning, recommendation systems, and medical domains.
These applications are slightly different from image classification or sentiment analyses and may require interpretations in a unique form, but most algorithms introduced previously can be used directly.

\subsubsection{Deep Reinforcement Learning (DRL) Related Domains}
Reinforcement learning (RL)~\citep{DBLP:journals/jair/KaelblingLM96} is an area of machine learning concerned with how intelligent agents ought to take actions in an environment in order to maximize the notion of cumulative reward.
Deep learning methods have recently enabled RL to decision-making problems that were previously intractable, such as playing games~\citep{DBLP:journals/nature/MnihKSRVBGRFOPB15,DBLP:journals/nature/SilverHMGSDSAPL16,DBLP:journals/nature/VinyalsBCMDCCPE19} and training robots~\citep{DBLP:journals/jmlr/LevineFDA16,DBLP:journals/ijrr/LevinePKIQ18,DBLP:journals/ijrr/OpenAI20}.
DRL is also applicable and shows potentials of application in healthcare, finance and business management~\citep{DBLP:journals/spm/ArulkumaranDBB17,DBLP:journals/corr/Li17b}, where human security and property safety issues should be considered, leading to the demands of explainable RL~\citep{DBLP:conf/cdmake/PuiuttaV20}.

According to recent surveys~\citep{DBLP:journals/spm/ArulkumaranDBB17,DBLP:journals/corr/Li17b}, DRL methods are generally based on DNNs to approximate value functions or find policies.
Most methods directly learn the objectives from raw inputs, especially for visual tasks where the images are used as inputs for estimating the value functions.
For those methods, input features related interpretation algorithms, such as LIME and SmoothGrad, have already been explored for explaining DRL methods~\cite{DBLP:conf/iclr/PuriVGKDK020,DBLP:conf/icml/GreydanusKDF18,DBLP:conf/aies/IyerLL0SS18,DBLP:conf/iclr/AtreyCJ20}.
However, as we discussed before, interpretation algorithms may expose different amount of information of the deep models, and in some real-world situations, different interpretation algorithms are required.
For critical problems concerning human security and property safety, showing the input-output relations of deep models is sometimes not persuadable for consumers.
The rationale inside the deep model may be required and has not been much investigated yet in this field.

\subsubsection{Recommendation Systems} 
The recommendation system~\citep{DBLP:reference/rsh/RicciRS11} is a subclass of the information retrieval domain that seeks to predict the ``rating'' or ``preference'' a user would give to an item.
With the growing information available on the Internet, it becomes more and more difficult to find the items of interest by users themselves.
For many web applications, the recommendation systems are an essential method for providing a better user experience~\citep{DBLP:journals/csur/ZhangYST19}.
Based on all kinds of information provided by users explicitly or implicitly, the recommendation system filters and sorts a list of items of interest in a personalization way.

There are three reasons for explainable recommendation systems.
The first one is to gain users' trust in the recommendation system.
Explanations help to improve the transparency, persuasiveness and user satisfaction of the recommendation system.
The second is to facilitate engineers to debug the recommendation algorithm.
Explanations provide analyses how the deep model works, and it would be easy to locate the bugs with explanations.
The first two arguments are borrowed from previous reviews~\citep{DBLP:journals/csur/ZhangYST19,DBLP:journals/ftir/ZhangC20}.
The third is to prevent the privacy and social issues.
The recommendations may be computed based on features that have privacy or ethical issues.
We would not like to have a recommendation system that may lead to these issues.
Explanations can thus be used to expose and prevent this problem.

Classic recommendation methods, including collaborate filtering~\citep{balabanovic1997fab,DBLP:conf/cscw/HerlockerKR00}, are interpretable, while the usages of black-box deep models~\cite{DBLP:journals/tmis/Gomez-UribeH16,DBLP:conf/recsys/CovingtonAS16,DBLP:conf/recsys/Cheng0HSCAACCIA16} increases the opacity of recommendation systems.
Recent works on explainable recommendation systems can be categorized following our proposed taxonomy, and most of them focus on designing interpretable modules~\cite{DBLP:conf/recsys/SeoHYL17,DBLP:conf/wsdm/TangW18,DBLP:conf/www/ChenZLM18,DBLP:conf/sigir/LiQPQDW19}.
We refer interested readers to the survey on explainable recommendation systems~\cite{DBLP:journals/ftir/ZhangC20}\footnote{We also note that whether the usage of deep models improves the recommendation system is an open discussion~\citep{DBLP:conf/recsys/DacremaCJ19}, but this is out of the scope of this survey.}.

\subsubsection{Deep Learning Applications to Medical Applications}
Deep learning methods have been recently applied on medical domains, especially on medical imaging analyses~\cite{DBLP:journals/mia/LitjensKBSCGLGS17}, such as the classification of Alzheimer’s~\cite{DBLP:journals/corr/abs-1905-00931}, lung cancer detection~\cite{hua2015computer}, tuberculosis diagnosis~\cite{rajpurkar2020chexaid}, retinal disease detection~\cite{DBLP:journals/artmed/SenguptaSLGL20}, etc.
Though researchers show the potentials of using deep learning methods in helping the diagnostics, the applications in the real-world situations of healthcare, clinics, hospitals and rehabilitation are very critical, because a single failure would cause irreparable damages.
Explanations for deep learning based methods are more urged in this field than in other fields, to gain the trust of physicians, regulators as well as the patients~\cite{DBLP:journals/jimaging/SinghSL20}.

Interpretation algorithms proposed in this specific domain have been surveyed~\cite{DBLP:journals/jimaging/SinghSL20,DBLP:journals/tnn/TjoaG21}.
Most of them are aligned with the general ones as reviewed in this work, because the network architectures are the same and the tasks are similar.
The difference mainly resides in the data distribution and the domain expert knowledge.
Interpretation algorithms are technically applicable and their trustworthiness can be evaluated in medical domains.
In spite of the advances, however, currently deep learning based methods have not achieved a significant deployment in the clinics still due to the lack of interpretability~\cite{DBLP:journals/jimaging/SinghSL20}.
This indicates that the new interpretation tools are still required in this domain.


\section{Trustworthiness Evaluations of Interpretation Algorithms and Model Interpretability Evaluations}
\label{sec:evaluation}

Previous section focuses on the interpretation algorithms and interpretation results. 
This section summarizes the current works in evaluating the trustworthiness of interpretation algorithms, and the deep models' interpretability.
To emphasize, the model interpretability is measured based on trustworthy interpretation algorithms.
Before introducing model interpretability evaluation, we present the evaluation methods for assuring the trustworthiness of interpretation algorithms in Section~\ref{subsec:algorithm-evaluation}.
Then, given a trustworthy interpretation algorithm, in Section~\ref{subsec:model-interpretability-evaluation} we present a few evaluation methods for the interpretability of deep models.

\subsection{Trustworthiness Evaluations of Interpretation Algorithm}
\label{subsec:algorithm-evaluation}

\paragraph{Perturbation-based Evaluations}
The perturbation-based evaluation of interpretation algorithms follows the intuition that flipping the most salient pixels first should lead to high performance decay.
Perturbation-based examples can therefore be used for the trustworthiness evaluations of interpretation algorithms~\cite{DBLP:journals/tnn/SamekBMLM17,DBLP:conf/iclr/HendrycksD19,DBLP:journals/corr/abs-1906-02032,croce2020robustbench}.
The main metric MoRF, Most Relevant First, (or LeRF, Least Relevant First, respectively), calculates the area under the curve (AUC), where the curve is of the probabilities predicted by the model after removing most relevant features (or least relevant features respectively).
MoRF would drop very quickly at beginning and LeRF would retain at a high value until the end, if the explanation is trustworthy.
They are usually used together and both have the same objective of evaluating the trustworthiness of the explanation.

In a different view~\cite{DBLP:conf/iccv/FongV17,DBLP:conf/nips/HookerEKK19} that \textit{``without re-training, it is unclear whether the degradation in model performance comes from the distribution shift or because the features that were removed are truly informative''}.
So Hooker et al.~\cite{DBLP:conf/nips/HookerEKK19} proposed to remove the most important features, extracted by the interpretation algorithms, and then retrain the model, to measure the degradation of model performance and evaluate the trustworthiness of interpretation algorithms.
We believe that the prohibitive computation cost added by the retraining step is meaningful for explaining the learning process (how the features/pixels were learned by a specific architecture of models), but contributes less to explain one trained model in a post-hoc way.

\paragraph{Evaluations by Randomizing Parameters}
There is no need for retraining in some cases, and we can identify untrustworthy interpretation algorithms by simply randomizing parameters.
Adebayo et al.~\cite{DBLP:conf/nips/AdebayoGMGHK18} found that even with random weights at the top layers of the network, a number of saliency map based approaches were still able to locate the important regions of the input images, and proved that these methods do not depend on the models.
Adebayo et al.~\cite{DBLP:conf/nips/AdebayoMLK20} summarized the uses of interpretation algorithms for model debugging, i.e., to detect spurious correlation artifacts (data contamination), diagnose mislabeled training examples (data contamination), differentiate between a (partially) re-initialized model and a trained one (model contamination), and detect out-of-distribution inputs (test-time contamination).

\paragraph{BAM}
Yang et al.~\cite{yang2019benchmarking} proposed a framework, named Benchmarking Attribution Methods (BAM), for benchmarking interpretation algorithms through a manually created dataset where objects are randomly pasted into images, and a set of models trained on that dataset.
BAM carefully generates a semi-natural dataset, where objects are copied into images of scenes, so each image has an object label and a scene label.
Then with models trained on this dataset and test examples, a target interpretation algorithm is evaluated by this framework, giving relative importance rankings for input features, which can be validated by ground truth from the generated dataset. 
The intuition behind BAM is that relative importance has a ground truth ranking, which can be controlled by the crafted dataset and used for comparing with the one given by interpretation methods, and then BAM can quantitatively evaluate the trustworthiness of the algorithm.

\paragraph{Trojaning}
Model trojaning attacks~\cite{DBLP:journals/corr/abs-1712-05526,DBLP:journals/corr/abs-1708-06733} indicate visual dataset contamination, where a subset of images are modified by giving a specific trigger (e.g., a yellow square is attached to the right bottom of the image) to the desired target.
This attack poisons the trained model that the trigger is the only feature for classifying the desired target.
Benefit from trojaning attacks, Lin et al.~\cite{DBLP:conf/kdd/LinLC21} proposed to verify the interpretation algorithm on the trojaned models.
The qualified algorithm should highlight pixels around the trigger in contaminated images instead of object parts.
Using the triggers as ground truth, Lin et al.~\cite{DBLP:conf/kdd/LinLC21} evaluated the trustworthiness of interpretation algorithms.

\paragraph{Infidelity and Sensitivity}
The desired properties relating to trustworthiness have been discussed in~\cite{DBLP:conf/iclr/AnconaCO018,DBLP:conf/nips/YehHSIR19}.
We reclaim the two definitions of (in)fidelity and sensitivity, which objectively and quantitatively measure the trustworthiness of interpretation algorithms.
Given a black-box function $\vf$, an interpretation algorithm $\Phi$, a random variable $\mI \in \sR^d$ with probability measure $\mu_\mI$, which represents meaningful perturbations of interest, and a given input neighborhood radius $r$, the infidelity and sensitivity of $\Phi$ of the target interpretation algorithm as:
\begin{align}
    &\text{INFD}(\Phi, \vf, \vx) = \E_{\mI \sim \mu_\mI} (\mI^T \Phi(\vf, \vx) - (\vf(\vx) - \vf(\vx - \mI))^2 ), \\
    &\text{SENS}_{\text{MAX}} = \max_{\|\vy - \vx\| \leq r}  \| \Phi(\vf, \vy) - \Phi(\vf, \vx) \|,
\end{align}
where $\mI$ represents significant perturbations around $\vx$, and can be specified in various ways.

\paragraph{ExpO Fidelity and Stability}
Plumb et al.~\citep{DBLP:conf/nips/PlumbACPXT20} proposed two metrics for measuring the desired properties of explanations and using them as regularization terms, to improve the explainability of trained models.
These two metrics can also be used as trustworthiness metrics for LIME and its variants, as they are able to evaluate the related fidelity and stability of proxy models.
We use ExpO-Fidelity and ExpO-Stability to refer the two metrics in this paragraph, where ExpO is short for Explanation-based Optimization, in order to avoid the confusion to the Infidelity and Sensitivity~\citep{DBLP:conf/nips/YehHSIR19}.
The formulas of ExpO-Fidelity and ExpO-Stability are 
\begin{align}
    &F(\vf, \vg, N_x) = \E_{x' \sim N_x} [(\vg(x') - \vf(x'))^2], \\
    &F(\vf, \ve, N_x) = \E_{x' \sim N_x} [ ||\ve(x, \vf) - \ve(x', \vf)||_2^2 ] \|,
\end{align}
where $\vg$ is the proxy model obtained by LIME or its variants, and $\ve(x, \vf)$ represents the post-hoc local explanation result given a local data point $x$ to explain and the model $\vf$.

\paragraph{Sensitivity to Hyperparameters}
Besides evaluations on the trustworthiness to the model, Bansal et al.~\cite{DBLP:conf/cvpr/BansalAN20a} proposed to measure the sensitivity to hyperparameters.
``\textit{It is important to carefully evaluate the pros and cons of interpretability methods with no hyperparameters and those that have.}''
In fact, the insensitivity to hyperparameters is also an important metric to trustworthiness.

\subsection{Model Interpretability Evaluation}
\label{subsec:model-interpretability-evaluation}

In some situations, different deep models exhibit different abilities to expose understandable terms to humans.
Even the same network architecture, training on different datasets may have different interpretability scores~\cite{DBLP:conf/cvpr/BauZKO017}.
Given the same trustworthy interpretation algorithm and any two models, model interpretability evaluation methods are used to measure and compare the interpretability between models.
In this subsection, we introduce four model interpretability evaluation methods, i.e., Network Dissection~\cite{DBLP:conf/cvpr/BauZKO017}, Pointing Game~\cite{DBLP:journals/ijcv/ZhangBLBSS18}, Consensus~\cite{DBLP:journals/corr/abs-2109-00707} and the one through OOD Samples~\cite{DBLP:conf/iclr/GeirhosRMBWB19,DBLP:conf/nips/GeirhosNMTBWB21}.



\begin{figure}
    \centering
    \subfloat[Image]{\includegraphics[width=0.18\linewidth]{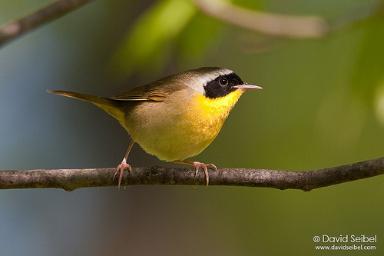}}\hspace{0.01\linewidth}
    \subfloat[\text{\footnotesize Human Label}]{\includegraphics[width=0.18\linewidth]{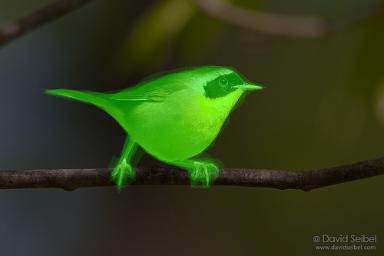}}\hspace{0.01\linewidth}
    \subfloat[LIME]{\includegraphics[width=0.18\linewidth]{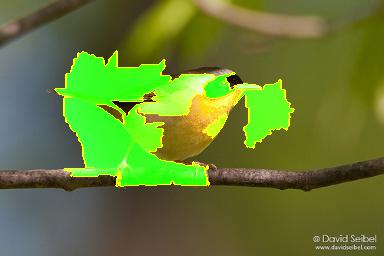}}\hspace{0.01\linewidth}
    \subfloat[GradCAM]{\includegraphics[width=0.18\linewidth]{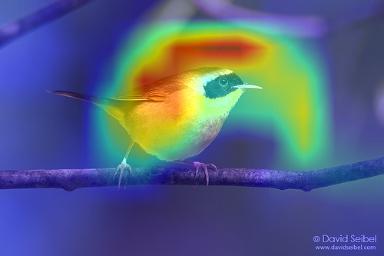}}\hspace{0.01\linewidth}
    \subfloat[SmoothGrad]{\includegraphics[width=0.18\linewidth]{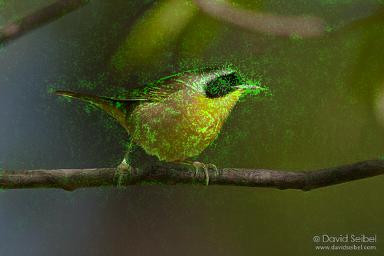}}\\
    \subfloat[Image]{\includegraphics[width=0.18\linewidth]{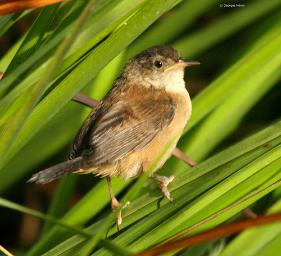}}\hspace{0.01\linewidth}
    \subfloat[\text{\footnotesize Human Label}]{\includegraphics[width=0.18\linewidth]{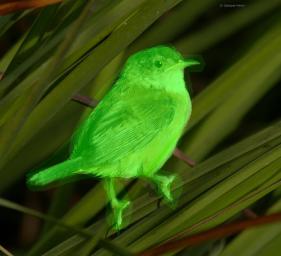}}\hspace{0.01\linewidth}
    \subfloat[LIME]{\includegraphics[width=0.18\linewidth]{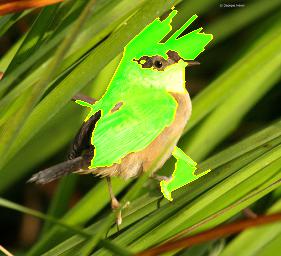}}\hspace{0.01\linewidth}
    \subfloat[GradCAM]{\includegraphics[width=0.18\linewidth]{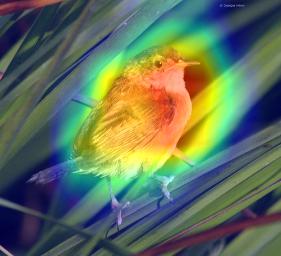}}\hspace{0.01\linewidth}
    \subfloat[SmoothGrad]{\includegraphics[width=0.18\linewidth]{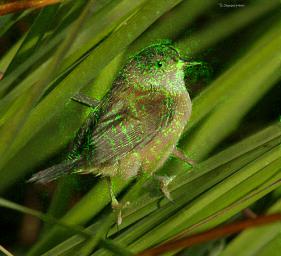}}\\
    \caption{Visualizations of semantic segmentation ground truth and interpretations from three popular algorithms, i.e. LIME, GradCAM and SmoothGrad, where the interpretation results are shown in different levels of granularity, i.e. superpixel, low-resolution, and pixel, respectively. We use the three algorithms to interpret images from CUB-200-2011~\cite{welinderetal2010caltech}, where the semantic segmentations are available.}
    
    \label{fig:visual_example}
\end{figure}

The basic idea for evaluating the model interpretability for Network Dissection~\cite{DBLP:conf/cvpr/BauZKO017}, Pointing Game~\cite{DBLP:journals/ijcv/ZhangBLBSS18} and Consensus~\cite{DBLP:journals/corr/abs-2109-00707} is to measure the overlap between semantic items (e.g., segmentation ground truth by humans, or cross-model ensemble of explanations) and interpretation results, as shown in Fig.~\ref{fig:visual_example}.

\paragraph{Network Dissection}
Network Dissection~\cite{DBLP:conf/cvpr/BauZKO017}, based on CAM~\cite{DBLP:conf/cvpr/ZhouKLOT16}, relies on a densely-labeled dataset where each image is labeled across colors, materials, textures, scenes, objects and object parts.
Given a CNN model, Network Dissection recovers the intermediate-layer feature maps used by the model for the classification, and then measures the mean intersection over union (mIoU) of each neuron between the activated locations with the labeled visual concepts.
A neuron is semantic if its mIoU is larger than a threshold.
Then the number of semantic neurons and the ratio are considered as the score for model interpretability.

\paragraph{Pointing Game}
The Pointing Game~\cite{DBLP:journals/ijcv/ZhangBLBSS18} measures the model interpretability via the localization accuracy.
This accuracy is equally the true positive rate between the computed explanation and the annotated object of interest.
It is similar to Network Dissection in the way that the pixel-wise or box-wise labels for visual concepts are required and the same intersection between explanations and annotations is measured. 

\paragraph{Consensus}
Consensus approach~\cite{DBLP:journals/corr/abs-2109-00707} incorporates an ensemble of deep models as a committee. 
Consensus first computes interpretations using a trustworthy interpretation algorithm (e.g., LIME~\cite{DBLP:conf/kdd/Ribeiro0G16}, SmoothGrad~\cite{DBLP:journals/corr/SmilkovTKVW17}) for every model in the committee, then obtains the consensus of interpretation from the entire committee through voting. 
Further, Consensus evaluates the interpretability of a model through matching its interpretation result (of LIME or SmoothGrad) to the consensus, and ranks the matching scores together with other deep models in the committee, so as to pursue the absolute and relative interpretability evaluation results.
Consensus uses LIME and SmoothGrad to validate its effectiveness, while Consensus is also compatible with other algorithms that interpret other targets, such as the rationale process, as long as the voting approach is suitable for the interpretation algorithm.

\paragraph{Through OOD Samples}
BAM~\cite{yang2019benchmarking} and Trojaning attacks~\cite{DBLP:journals/corr/abs-1712-05526,DBLP:journals/corr/abs-1708-06733} create datasets that are different from natural distributions, and train the models on such datasets.
Models trained on such datasets are used to verify the trustworthiness of interpretation algorithms because they should suffer from the attacks on the datasets.
In another way, one can use the such ideas of out-of-distribution (OOD) samples to directly evaluate the deep models where the OOD samples were not seen during training.
\cite{DBLP:conf/iclr/GeirhosRMBWB19,DBLP:conf/nips/GeirhosNMTBWB21} generated different OOD datasets and tested with classic deep models and human observers to record the errors that they made on these datasets.
With sophisticated designs of datasets and experiments, they found that the consistency between humans and deep models is closing.
These evaluations show the interpretability of deep models in a general way, to present that the visual recognition of models is partially consistent with humans'.
This could be easily extended to the comparison within models.



\subsection{Human-Centered/User-Study Evaluations}

User studies involving humans are a commonly used method for evaluating the trustworthiness of interpretation algorithms and model interpretability.
We combine these two directions and introduce them here, as the designed user-study experiments may be capable of performing the two evaluations simultaneously.

An approach to evaluate the algorithm of counterfactual examples~\cite{DBLP:conf/iclr/AntoranBAWH21} was proposed, where a user-study experiment was used to validate their approaches.
This user-study experiment aims at verifying whether humans can predict the deep model's decision.
Specifically, several (clean and counterfactual) samples with models' predictions are presented to users, and then a new sample is shown to ask the user if the model can make the correct decisions or not.
Another approach based on decision trees and sets, designs descriptive and multiple-choice questions to test the user’s understanding of the decision boundaries of the classes in the data, in order to evaluate the interpretability of their proposed Bayesian Decision Lists.
\cite{DBLP:journals/corr/abs-1902-03501} designed the user-study experiments following the idea that interpretability is the user's ability to predict the model's changes in response to changes in input.
More user studies can be found in~\cite{DBLP:conf/www/Grgic-HlacaRGW18,DBLP:journals/corr/abs-1902-00006,DBLP:conf/flairs/IslamEG20}.



\subsection{Concluding Remarks}

\begin{table}[t]
\renewcommand{\arraystretch}{1.25}
\caption{List of evaluation methods. There are two categories of evaluations as introduced in this work: Trustworthiness Evaluations of Interpretation Algorithm (T.E.I.A) and Model Interpretability Evaluation (M.I.E), with respect to Section~\ref{subsec:algorithm-evaluation} and Section~\ref{subsec:model-interpretability-evaluation}. Additional notes are added as a description for the speciality of the evaluation method.}
\centering
\begin{tabular}{|l|l|l|}
\hline
\multicolumn{1}{|l|}{\textbf{Method   Name}} & \multicolumn{1}{c|}{\textbf{Category}} & \multicolumn{1}{c|}{\textbf{Additional Notes}}       \\ \hline
Perturbation                                 & T.E.I.A                                & AUC scores of MoRF, LeRF                             \\ \hline
Randoming Parameters                         & T.E.I.A                                & Filtering Irrelavant Algorithms                      \\ \hline
BAM                                          & T.E.I.A                                & Based on a semi-natural dataset                      \\ \hline
Trojaning                                    & T.E.I.A                                & Based on a semi-natural dataset                      \\ \hline
Infidelity and Sensitivity                   & T.E.I.A                                & -                 \\ \hline
Expo Fidelity and Stability                   & T.E.I.A                                & Available only for LIME and variants                 \\ \hline
Sensitivity to Hyperparameters               & T.E.I.A                                & -                                                    \\ \hline
Network Dissection                           & M.I.E                                  & Based on a densely labeled dataset                   \\ \hline
Pointing Game                           & M.I.E                                  & Requires pixel-wise or box-wise labels                   \\ \hline
Consensus                                    & M.I.E                                  & Based on cross-model explanations \\ \hline
Through OOD Samples                          & M.I.E                                  & Based on OOD datasets                                \\ \hline
\end{tabular}
\label{table:evaluations}
\end{table}

We summarize the evaluation methods in Table~\ref{table:evaluations}.
We have to note that assessing the trustworthiness of interpretation algorithms is challenging.
While a small number of algorithms benefit from intrinsic properties of deep models, e.g., closed-form interpretations, the trustworthiness of most algorithms remains to be evaluated.
Despite filtering approaches (such as randomizing the weights~\cite{DBLP:conf/nips/AdebayoGMGHK18}) to picking out irrelevant interpretation algorithms, reasonable and practical evaluation approaches for directly assessing the trustworthiness are also reviewed.
Given a trustworthy algorithm, the interpretability can be evaluated between models, to compare the degree of being understandable.
If the algorithm is not trustworthy, it does not make sense to compare the interpretability of models using unreliable interpretation results.
A few model interpretability evaluation methods are introduced, while more model interpretability evaluations should be explored in the future.
We also note that subjective human-centered user studies are one important evaluation tool that can be used for evaluating both interpretation algorithms and model interpretability, thanks to the flexibility of designing arbitrary experiments for various objectives.

\section{Impact beyond Interpretations}
\label{sec:relations}


Deep models have many unknown phenomenon and properties, e.g., adversarial attacks, memorization capacity, generalization ability etc.
(Lack of) interpretation and (low) interpretability are one of them.
Interestingly, besides the original motivations for explaining black-box deep models, interpretations related terms have been connected to existing findings about deep models.
In this section, we present two fields that are widely known to be related to interpretations.

\subsection{Interpretability, Adversarial Attacks and Robustness}
Recent studies on adversarial examples have found positive connections between model interpretability and adversarial robustness. 
Two teams~\cite{DBLP:conf/aaai/RossD18,DBLP:conf/iclr/TsiprasSETM19} first observed that compared to standard models, adversarially trained models show more interpretable input gradients. 
Etmann et al.~\cite{DBLP:conf/icml/EtmannLMS19} theoretically proved that the increase in adversarial robustness improves the alignment between input and its respective input gradient, using the case of a linear binary classifier.
Zhang et al.~\cite{DBLP:conf/icml/ZhangZ19} further analyzed how adversarially trained models achieve robustness from an interpretation perspective, showing that adversarially robust models rely on fewer texture features and are more shape-biased, which is regarded as coincide more with the human interpretation. 
Essentially, the connection between adversarial examples and gradient-based interpretations may come from their common dependence on the input gradient.

For future works, these observations could (1) motivate new understandings about how deep models work and (2) explore the connections between interpretation related terms and other properties of deep models.


\subsection{Learning from Interpretations}
As containing rich information about the location of discriminative features, interpretation results can also be utilized to guide training strategies such as data augmentations and regularization approaches, especially for vision tasks. 
For example, Kim et al.~\cite{DBLP:conf/icml/KimCS20} proposed to improve Mixup~\cite{DBLP:conf/iclr/ZhangCDL18} by leveraging the saliency map~\cite{DBLP:journals/corr/SimonyanVZ13}. 
Specifically, they aimed to seek an optimal transport that maximizes the exposed saliency. 
Zagoruyko et al.~\cite{DBLP:conf/iclr/ZagoruykoK17} imposed the regularizer to encourage the alignment of saliency maps between the teacher and student networks for effective knowledge distillation. 
Wickramanayake et al.~\cite{DBLP:conf/nips/WickramanayakeH21} also used interpretations to generate efficient augmented data samples to train the model, for improving the interpretability and the model performance.
Interpretations sometimes can be used as weak labels in specific tasks. 
For example, Lai et al.~\cite{DBLP:conf/ijcai/LaiG17} introduced a saliency-guided learning approach for weakly supervised object detection.
Many weakly object localization and weakly semantic segmentation methods~\cite{DBLP:conf/cvpr/AhnK18,DBLP:conf/icml/KimCS20,DBLP:conf/cvpr/Yao0XZS0T021} start from an interpretation, and obtain promising results.

From these works, we believe that the interpretability and model performance are not two contradictory measures and that they can be improved simultaneously.
Future works could further focus on this direction.





\section{Open-Source Libraries for Deep Learning Interpretation}
\label{sec:library}
To simplify future researches and practical usages, we introduce several open-source libraries that implement popular interpretation algorithms based on mainstream deep learning frameworks, such as TF-Explainer\footnote{\url{https://github.com/sicara/tf-explain}} based on Tensorflow~\cite{tensorflow2015-whitepaper}, Captum\footnote{\url{https://github.com/pytorch/captum}} based on PyTorch~\cite{DBLP:conf/nips/PaszkeGMLBCKLGA19} and InterpretDL\footnote{\url{https://github.com/PaddlePaddle/InterpretDL}} based on PaddlePaddle~\cite{ma2019paddlepaddle}.
Note that TF-explainer and Captum mainly include algorithms that target at features with gradient-based techniques.
Some other popular libraries focus on machine learning and have not involved deep models, such as interpretml\footnote{\url{https://github.com/interpretml/interpret}}, AIX360\footnote{\url{https://github.com/Trusted-AI/AIX360}} etc., and the library LIT\footnote{\url{https://github.com/PAIR-code/lit}} that is for NLP models. 

\section{Discussions and Conclusions}

In this paper, we review the recent research on interpretation algorithms, model interpretability, and the connections to other deep learning factors.

First of all, to address the research efforts in interpretations, we clarify the main concepts of interpretation algorithms and model interpretability that were usually confused, and connect them by introducing the notion of trustworthiness of interpretation algorithms.

Second, we propose a new taxonomy and elaborate the design of several recent interpretation algorithms, from different perspectives according to the proposed taxonomy. 
Our work reviews the recent advances in interpretation algorithms, and provides a clear categorization, to help future researches to better compare new algorithms with the most related works, or progress in unexplored directions. 

Third, we survey the performance metrics for evaluating the trustworthiness of interpretation algorithms, to guarantee the appropriate usages of the interpretation results. 
These metrics can be used to quantitatively compare between the interpretation algorithms.
The proposition of new algorithms can be supported by comparing these metrics with related works, instead of by providing tenuous descriptions and qualitative visualizations.

Further, we summarize the current work in evaluating models' interpretability given trustworthy interpretation algorithms. 
Based on these evaluations, more relations between interpretability and other metrics could be found for deep models, possibly leading to further understandings about the deep learning.
However, there are not many evaluation methods for measuring the interpretability, though the existing ones are largely aligned for popular network architectures.
Designing new methods of evaluating models' interpretability could be one of the important research directions.

Finally, we review and discuss the connections between deep models' interpretations and other factors, such as adversarial robustness and learning from interpretations.
New understandings how deep models could be observed and analyzed.
Note that many interpretation algorithms and evaluation approaches are open-sourced and there are some useful libraries to simplify the practical usages and future researches. 

{\small 
\bibliographystyle{plain}
\bibliography{reference}

\begin{thebibliography}{100}

\bibitem{tensorflow2015-whitepaper}
Mart\'{\i}n Abadi, Ashish Agarwal, Paul Barham, Eugene Brevdo, Zhifeng Chen,
  Craig Citro, Greg~S. Corrado, Andy Davis, Jeffrey Dean, Matthieu Devin,
  Sanjay Ghemawat, Ian Goodfellow, Andrew Harp, Geoffrey Irving, Michael Isard,
  Yangqing Jia, Rafal Jozefowicz, Lukasz Kaiser, Manjunath Kudlur, Josh
  Levenberg, Dandelion Man\'{e}, Rajat Monga, Sherry Moore, Derek Murray, Chris
  Olah, Mike Schuster, Jonathon Shlens, Benoit Steiner, Ilya Sutskever, Kunal
  Talwar, Paul Tucker, Vincent Vanhoucke, Vijay Vasudevan, Fernanda Vi\'{e}gas,
  Oriol Vinyals, Pete Warden, Martin Wattenberg, Martin Wicke, Yuan Yu, and
  Xiaoqiang Zheng.
\newblock {TensorFlow}: Large-scale machine learning on heterogeneous systems,
  2015.
\newblock Software available from tensorflow.org.

\bibitem{DBLP:conf/acl/AbnarZ20}
Samira Abnar and Willem~H. Zuidema.
\newblock Quantifying attention flow in transformers.
\newblock In Dan Jurafsky, Joyce Chai, Natalie Schluter, and Joel~R. Tetreault,
  editors, {\em Proceedings of the 58th Annual Meeting of the Association for
  Computational Linguistics, {ACL}}. Association for Computational Linguistics,
  2020.

\bibitem{DBLP:conf/nips/AdebayoGMGHK18}
Julius Adebayo, Justin Gilmer, Michael Muelly, Ian~J. Goodfellow, Moritz Hardt,
  and Been Kim.
\newblock Sanity checks for saliency maps.
\newblock In Samy Bengio, Hanna~M. Wallach, Hugo Larochelle, Kristen Grauman,
  Nicol{\`{o}} Cesa{-}Bianchi, and Roman Garnett, editors, {\em Advances in
  Neural Information Processing Systems 31: Annual Conference on Neural
  Information Processing Systems 2018, NeurIPS 2018, December 3-8, 2018,
  Montr{\'{e}}al, Canada}, pages 9525--9536, 2018.

\bibitem{DBLP:conf/nips/AdebayoMLK20}
Julius Adebayo, Michael Muelly, Ilaria Liccardi, and Been Kim.
\newblock Debugging tests for model explanations.
\newblock In Hugo Larochelle, Marc'Aurelio Ranzato, Raia Hadsell,
  Maria{-}Florina Balcan, and Hsuan{-}Tien Lin, editors, {\em Advances in
  Neural Information Processing Systems 33: Annual Conference on Neural
  Information Processing Systems 2020, NeurIPS 2020, December 6-12, 2020,
  virtual}, 2020.

\bibitem{DBLP:conf/nips/AgarwalMFZLCH21}
Rishabh Agarwal, Levi Melnick, Nicholas Frosst, Xuezhou Zhang, Benjamin~J.
  Lengerich, Rich Caruana, and Geoffrey~E. Hinton.
\newblock Neural additive models: Interpretable machine learning with neural
  nets.
\newblock In Marc'Aurelio Ranzato, Alina Beygelzimer, Yann~N. Dauphin, Percy
  Liang, and Jennifer~Wortman Vaughan, editors, {\em Advances in Neural
  Information Processing Systems 34: Annual Conference on Neural Information
  Processing Systems 2021, NeurIPS 2021, December 6-14, 2021, virtual}, pages
  4699--4711, 2021.

\bibitem{DBLP:journals/corr/abs-1909-04200}
Isaac Ahern, Adam Noack, Luis Guzman{-}Nateras, Dejing Dou, Boyang Li, and Jun
  Huan.
\newblock Normlime: {A} new feature importance metric for explaining deep
  neural networks.
\newblock {\em CoRR}, abs/1909.04200, 2019.

\bibitem{DBLP:conf/cvpr/AhnK18}
Jiwoon Ahn and Suha Kwak.
\newblock Learning pixel-level semantic affinity with image-level supervision
  for weakly supervised semantic segmentation.
\newblock In {\em 2018 {IEEE} Conference on Computer Vision and Pattern
  Recognition, {CVPR} 2018, Salt Lake City, UT, USA, June 18-22, 2018}, pages
  4981--4990. Computer Vision Foundation / {IEEE} Computer Society, 2018.

\bibitem{DBLP:journals/corr/abs-1806-08049}
David Alvarez{-}Melis and Tommi~S. Jaakkola.
\newblock On the robustness of interpretability methods.
\newblock {\em CoRR}, abs/1806.08049, 2018.

\bibitem{DBLP:conf/iclr/AnconaCO018}
Marco Ancona, Enea Ceolini, Cengiz {\"{O}}ztireli, and Markus Gross.
\newblock Towards better understanding of gradient-based attribution methods
  for deep neural networks.
\newblock In {\em 6th International Conference on Learning Representations,
  {ICLR} 2018, Vancouver, BC, Canada, April 30 - May 3, 2018, Conference Track
  Proceedings}. OpenReview.net, 2018.

\bibitem{DBLP:journals/ijrr/OpenAI20}
Marcin Andrychowicz, Bowen Baker, Maciek Chociej, Rafal J{\'{o}}zefowicz, Bob
  McGrew, Jakub Pachocki, Arthur Petron, Matthias Plappert, Glenn Powell, Alex
  Ray, Jonas Schneider, Szymon Sidor, Josh Tobin, Peter Welinder, Lilian Weng,
  and Wojciech Zaremba.
\newblock Learning dexterous in-hand manipulation.
\newblock {\em Int. J. Robotics Res.}, 39(1), 2020.

\bibitem{DBLP:conf/iclr/AntoranBAWH21}
Javier Antor{\'{a}}n, Umang Bhatt, Tameem Adel, Adrian Weller, and
  Jos{\'{e}}~Miguel Hern{\'{a}}ndez{-}Lobato.
\newblock Getting a {CLUE:} {A} method for explaining uncertainty estimates.
\newblock In {\em 9th International Conference on Learning Representations,
  {ICLR} 2021, Virtual Event, Austria, May 3-7, 2021}. OpenReview.net, 2021.

\bibitem{DBLP:journals/corr/abs-1911-07749}
Andr{\'{e}} Artelt and Barbara Hammer.
\newblock On the computation of counterfactual explanations - {A} survey.
\newblock {\em CoRR}, abs/1911.07749, 2019.

\bibitem{DBLP:journals/spm/ArulkumaranDBB17}
Kai Arulkumaran, Marc~Peter Deisenroth, Miles Brundage, and Anil~Anthony
  Bharath.
\newblock Deep reinforcement learning: {A} brief survey.
\newblock {\em {IEEE} Signal Process. Mag.}, 34(6):26--38, 2017.

\bibitem{DBLP:conf/acl/AtanasovaSLA20}
Pepa Atanasova, Jakob~Grue Simonsen, Christina Lioma, and Isabelle Augenstein.
\newblock Generating fact checking explanations.
\newblock In Dan Jurafsky, Joyce Chai, Natalie Schluter, and Joel~R. Tetreault,
  editors, {\em Proceedings of the 58th Annual Meeting of the Association for
  Computational Linguistics, {ACL}}. Association for Computational Linguistics,
  2020.

\bibitem{DBLP:conf/iclr/AtreyCJ20}
Akanksha Atrey, Kaleigh Clary, and David~D. Jensen.
\newblock Exploratory not explanatory: Counterfactual analysis of saliency maps
  for deep reinforcement learning.
\newblock In {\em 8th International Conference on Learning Representations,
  {ICLR} 2020, Addis Ababa, Ethiopia, April 26-30, 2020}. OpenReview.net, 2020.

\bibitem{bach2015pixel}
Sebastian Bach, Alexander Binder, Gr{\'e}goire Montavon, Frederick Klauschen,
  Klaus-Robert M{\"u}ller, and Wojciech Samek.
\newblock On pixel-wise explanations for non-linear classifier decisions by
  layer-wise relevance propagation.
\newblock {\em PloS one}, 2015.

\bibitem{DBLP:journals/corr/BahdanauCB14}
Dzmitry Bahdanau, Kyunghyun Cho, and Yoshua Bengio.
\newblock Neural machine translation by jointly learning to align and
  translate.
\newblock In Yoshua Bengio and Yann LeCun, editors, {\em International
  Conference on Learning Representations}, 2015.

\bibitem{DBLP:conf/nips/BajajCXPWLZ21}
Mohit Bajaj, Lingyang Chu, Zi~Yu Xue, Jian Pei, Lanjun Wang, Peter~Cho{-}Ho
  Lam, and Yong Zhang.
\newblock Robust counterfactual explanations on graph neural networks.
\newblock In Marc'Aurelio Ranzato, Alina Beygelzimer, Yann~N. Dauphin, Percy
  Liang, and Jennifer~Wortman Vaughan, editors, {\em Advances in Neural
  Information Processing Systems 34: Annual Conference on Neural Information
  Processing Systems 2021, NeurIPS 2021, December 6-14, 2021, virtual}, pages
  5644--5655, 2021.

\bibitem{balabanovic1997fab}
Marko Balabanovi{\'c} and Yoav Shoham.
\newblock Fab: content-based, collaborative recommendation.
\newblock {\em Communications of the ACM}, 1997.

\bibitem{DBLP:journals/corr/abs-1905-13686}
Federico Baldassarre and Hossein Azizpour.
\newblock Explainability techniques for graph convolutional networks.
\newblock {\em CoRR}, abs/1905.13686, 2019.

\bibitem{DBLP:conf/nips/BaldockMN21}
Robert J.~N. Baldock, Hartmut Maennel, and Behnam Neyshabur.
\newblock Deep learning through the lens of example difficulty.
\newblock In Marc'Aurelio Ranzato, Alina Beygelzimer, Yann~N. Dauphin, Percy
  Liang, and Jennifer~Wortman Vaughan, editors, {\em Advances in Neural
  Information Processing Systems 34: Annual Conference on Neural Information
  Processing Systems 2021, NeurIPS 2021, December 6-14, 2021, virtual}, pages
  10876--10889, 2021.

\bibitem{DBLP:conf/cvpr/BansalAN20a}
Naman Bansal, Chirag Agarwal, and Anh Nguyen.
\newblock {SAM:} the sensitivity of attribution methods to hyperparameters.
\newblock In {\em 2020 {IEEE/CVF} Conference on Computer Vision and Pattern
  Recognition, {CVPR} 2020, Seattle, WA, USA, June 13-19, 2020}, pages
  8670--8680. Computer Vision Foundation / {IEEE}, 2020.

\bibitem{DBLP:conf/pkdd/BarbalauCIP20}
Antonio Barbalau, Adrian Cosma, Radu~Tudor Ionescu, and Marius Popescu.
\newblock A generic and model-agnostic exemplar synthetization framework for
  explainable {AI}.
\newblock In Frank Hutter, Kristian Kersting, Jefrey Lijffijt, and Isabel
  Valera, editors, {\em Machine Learning and Knowledge Discovery in Databases -
  European Conference, {ECML} {PKDD} 2020, Ghent, Belgium, September 14-18,
  2020, Proceedings, Part {II}}, volume 12458 of {\em Lecture Notes in Computer
  Science}, pages 190--205. Springer, 2020.

\bibitem{DBLP:conf/cvpr/BauZKO017}
David Bau, Bolei Zhou, Aditya Khosla, Aude Oliva, and Antonio Torralba.
\newblock Network dissection: Quantifying interpretability of deep visual
  representations.
\newblock In {\em 2017 {IEEE} Conference on Computer Vision and Pattern
  Recognition, {CVPR} 2017, Honolulu, HI, USA, July 21-26, 2017}, pages
  3319--3327. {IEEE} Computer Society, 2017.

\bibitem{DBLP:conf/iclr/BauZSZTFT19}
David Bau, Jun{-}Yan Zhu, Hendrik Strobelt, Bolei Zhou, Joshua~B. Tenenbaum,
  William~T. Freeman, and Antonio Torralba.
\newblock {GAN} dissection: Visualizing and understanding generative
  adversarial networks.
\newblock In {\em 7th International Conference on Learning Representations,
  {ICLR} 2019, New Orleans, LA, USA, May 6-9, 2019}. OpenReview.net, 2019.

\bibitem{bien2011prototype}
Jacob Bien and Robert Tibshirani.
\newblock Prototype selection for interpretable classification.
\newblock {\em The Annals of Applied Statistics}, 2011.

\bibitem{DBLP:conf/icann/BinderMLMS16}
Alexander Binder, Gr{\'{e}}goire Montavon, Sebastian Lapuschkin, Klaus{-}Robert
  M{\"{u}}ller, and Wojciech Samek.
\newblock Layer-wise relevance propagation for neural networks with local
  renormalization layers.
\newblock In Alessandro E.~P. Villa, Paolo Masulli, and Antonio Javier~Pons
  Rivero, editors, {\em Artificial Neural Networks and Machine Learning -
  {ICANN} 2016 - 25th International Conference on Artificial Neural Networks,
  Barcelona, Spain, September 6-9, 2016, Proceedings, Part {II}}, volume 9887
  of {\em Lecture Notes in Computer Science}, pages 63--71. Springer, 2016.

\bibitem{DBLP:journals/corr/abs-1910-13427}
Nicholas Carlini, {\'{U}}lfar Erlingsson, and Nicolas Papernot.
\newblock Distribution density, tails, and outliers in machine learning:
  Metrics and applications.
\newblock {\em CoRR}, abs/1910.13427, 2019.

\bibitem{carvalho2019machine}
Diogo~V Carvalho, Eduardo~M Pereira, and Jaime~S Cardoso.
\newblock Machine learning interpretability: A survey on methods and metrics.
\newblock {\em Electronics}, 2019.

\bibitem{DBLP:journals/corr/abs-1810-00069}
Anirban Chakraborty, Manaar Alam, Vishal Dey, Anupam Chattopadhyay, and Debdeep
  Mukhopadhyay.
\newblock Adversarial attacks and defences: {A} survey.
\newblock {\em CoRR}, abs/1810.00069, 2018.

\bibitem{DBLP:conf/iclr/ChangCGD19}
Chun{-}Hao Chang, Elliot Creager, Anna Goldenberg, and David Duvenaud.
\newblock Explaining image classifiers by counterfactual generation.
\newblock In {\em 7th International Conference on Learning Representations,
  {ICLR} 2019, New Orleans, LA, USA, May 6-9, 2019}. OpenReview.net, 2019.

\bibitem{DBLP:conf/wacv/ChattopadhyaySH18}
Aditya Chattopadhyay, Anirban Sarkar, Prantik Howlader, and Vineeth~N.
  Balasubramanian.
\newblock {Grad-CAM}++: Generalized gradient-based visual explanations for deep
  convolutional networks.
\newblock In {\em 2018 {IEEE} Winter Conference on Applications of Computer
  Vision, {WACV} 2018, Lake Tahoe, NV, USA, March 12-15, 2018}, pages 839--847.
  {IEEE} Computer Society, 2018.

\bibitem{DBLP:conf/iccv/CheferGW21}
Hila Chefer, Shir Gur, and Lior Wolf.
\newblock Generic attention-model explainability for interpreting bi-modal and
  encoder-decoder transformers.
\newblock In {\em 2021 {IEEE/CVF} International Conference on Computer Vision,
  {ICCV} 2021, Montreal, QC, Canada, October 10-17, 2021}, pages 387--396.
  {IEEE}, 2021.

\bibitem{DBLP:conf/cvpr/CheferGW21}
Hila Chefer, Shir Gur, and Lior Wolf.
\newblock Transformer interpretability beyond attention visualization.
\newblock In {\em {IEEE} Conference on Computer Vision and Pattern Recognition,
  {CVPR} 2021, virtual, June 19-25, 2021}, pages 782--791. Computer Vision
  Foundation / {IEEE}, 2021.

\bibitem{DBLP:conf/nips/ChenLTBRS19}
Chaofan Chen, Oscar Li, Daniel Tao, Alina Barnett, Cynthia Rudin, and Jonathan
  Su.
\newblock This looks like that: Deep learning for interpretable image
  recognition.
\newblock In Hanna~M. Wallach, Hugo Larochelle, Alina Beygelzimer, Florence
  d'Alch{\'{e}}{-}Buc, Emily~B. Fox, and Roman Garnett, editors, {\em Advances
  in Neural Information Processing Systems 32: Annual Conference on Neural
  Information Processing Systems 2019, NeurIPS 2019, December 8-14, 2019,
  Vancouver, BC, Canada}, pages 8928--8939, 2019.

\bibitem{DBLP:conf/www/ChenZLM18}
Chong Chen, Min Zhang, Yiqun Liu, and Shaoping Ma.
\newblock Neural attentional rating regression with review-level explanations.
\newblock In Pierre{-}Antoine Champin, Fabien Gandon, Mounia Lalmas, and
  Panagiotis~G. Ipeirotis, editors, {\em Proceedings of the 2018 World Wide Web
  Conference on World Wide Web, {WWW} 2018, Lyon, France, April 23-27, 2018},
  pages 1583--1592. {ACM}, 2018.

\bibitem{DBLP:journals/corr/abs-1712-05526}
Xinyun Chen, Chang Liu, Bo~Li, Kimberly Lu, and Dawn Song.
\newblock Targeted backdoor attacks on deep learning systems using data
  poisoning.
\newblock {\em CoRR}, abs/1712.05526, 2017.

\bibitem{DBLP:conf/aaai/ChenLYWM21}
Yuanyuan Chen, Boyang Li, Han Yu, Pengcheng Wu, and Chunyan Miao.
\newblock Hydra: Hypergradient data relevance analysis for interpreting deep
  neural networks.
\newblock In {\em Thirty-Fifth {AAAI} Conference on Artificial Intelligence,
  {AAAI} 2021, Thirty-Third Conference on Innovative Applications of Artificial
  Intelligence, {IAAI} 2021, The Eleventh Symposium on Educational Advances in
  Artificial Intelligence, {EAAI} 2021, Virtual Event, February 2-9, 2021},
  pages 7081--7089. {AAAI} Press, 2021.

\bibitem{DBLP:conf/recsys/Cheng0HSCAACCIA16}
Heng{-}Tze Cheng, Levent Koc, Jeremiah Harmsen, Tal Shaked, Tushar Chandra,
  Hrishi Aradhye, Glen Anderson, Greg Corrado, Wei Chai, Mustafa Ispir, Rohan
  Anil, Zakaria Haque, Lichan Hong, Vihan Jain, Xiaobing Liu, and Hemal Shah.
\newblock Wide {\&} deep learning for recommender systems.
\newblock In Alexandros Karatzoglou, Bal{\'{a}}zs Hidasi, Domonkos Tikk,
  Oren~Sar Shalom, Haggai Roitman, Bracha Shapira, and Lior Rokach, editors,
  {\em Proceedings of the 1st Workshop on Deep Learning for Recommender
  Systems, DLRS@RecSys 2016, Boston, MA, USA, September 15, 2016}, pages 7--10.
  {ACM}, 2016.

\bibitem{DBLP:conf/recsys/CovingtonAS16}
Paul Covington, Jay Adams, and Emre Sargin.
\newblock Deep neural networks for youtube recommendations.
\newblock In Shilad Sen, Werner Geyer, Jill Freyne, and Pablo Castells,
  editors, {\em Proceedings of the 10th {ACM} Conference on Recommender
  Systems, Boston, MA, USA, September 15-19, 2016}, pages 191--198. {ACM},
  2016.

\bibitem{croce2020robustbench}
Francesco Croce, Maksym Andriushchenko, Vikash Sehwag, Edoardo Debenedetti,
  Nicolas Flammarion, Mung Chiang, Prateek Mittal, and Matthias Hein.
\newblock Robustbench: a standardized adversarial robustness benchmark.
\newblock {\em arXiv preprint arXiv:2010.09670}, 2020.

\bibitem{DBLP:conf/recsys/DacremaCJ19}
Maurizio~Ferrari Dacrema, Paolo Cremonesi, and Dietmar Jannach.
\newblock Are we really making much progress? {A} worrying analysis of recent
  neural recommendation approaches.
\newblock In Toine Bogers, Alan Said, Peter Brusilovsky, and Domonkos Tikk,
  editors, {\em Proceedings of the 13th {ACM} Conference on Recommender
  Systems, RecSys 2019, Copenhagen, Denmark, September 16-20, 2019}, pages
  101--109. {ACM}, 2019.

\bibitem{DBLP:conf/cvpr/DengDSLL009}
Jia Deng, Wei Dong, Richard Socher, Li{-}Jia Li, Kai Li, and Li~Fei{-}Fei.
\newblock Imagenet: {A} large-scale hierarchical image database.
\newblock In {\em 2009 {IEEE} Computer Society Conference on Computer Vision
  and Pattern Recognition {(CVPR} 2009), 20-25 June 2009, Miami, Florida,
  {USA}}, pages 248--255. {IEEE} Computer Society, 2009.

\bibitem{DBLP:conf/wacv/DesaiR20}
Saurabh Desai and Harish~G. Ramaswamy.
\newblock {Ablation-CAM}: Visual explanations for deep convolutional network
  via gradient-free localization.
\newblock In {\em {IEEE} Winter Conference on Applications of Computer Vision,
  {WACV} 2020, Snowmass Village, CO, USA, March 1-5, 2020}, pages 972--980.
  {IEEE}, 2020.

\bibitem{DBLP:conf/naacl/DevlinCLT19}
Jacob Devlin, Ming{-}Wei Chang, Kenton Lee, and Kristina Toutanova.
\newblock {BERT:} pre-training of deep bidirectional transformers for language
  understanding.
\newblock In Jill Burstein, Christy Doran, and Thamar Solorio, editors, {\em
  Proceedings of the 2019 Conference of the North American Chapter of the
  Association for Computational Linguistics: Human Language Technologies,
  {NAACL-HLT} 2019, Minneapolis, MN, USA, June 2-7, 2019, Volume 1 (Long and
  Short Papers)}, pages 4171--4186. Association for Computational Linguistics,
  2019.

\bibitem{DBLP:conf/cvpr/DongSZZ17}
Yinpeng Dong, Hang Su, Jun Zhu, and Bo~Zhang.
\newblock Improving interpretability of deep neural networks with semantic
  information.
\newblock In {\em 2017 {IEEE} Conference on Computer Vision and Pattern
  Recognition, {CVPR} 2017, Honolulu, HI, USA, July 21-26, 2017}, pages
  975--983. {IEEE} Computer Society, 2017.

\bibitem{doshi2017towards}
Finale Doshi-Velez and Been Kim.
\newblock Towards a rigorous science of interpretable machine learning.
\newblock {\em arXiv preprint arXiv:1702.08608}, 2017.

\bibitem{DBLP:conf/iclr/DosovitskiyB0WZ21}
Alexey Dosovitskiy, Lucas Beyer, Alexander Kolesnikov, Dirk Weissenborn,
  Xiaohua Zhai, Thomas Unterthiner, Mostafa Dehghani, Matthias Minderer, Georg
  Heigold, Sylvain Gelly, Jakob Uszkoreit, and Neil Houlsby.
\newblock An image is worth 16x16 words: Transformers for image recognition at
  scale.
\newblock In {\em 9th International Conference on Learning Representations,
  {ICLR} 2021, Virtual Event, Austria, May 3-7, 2021}. OpenReview.net, 2021.

\bibitem{erhan2009visualizing}
Dumitru Erhan, Yoshua Bengio, Aaron Courville, and Pascal Vincent.
\newblock Visualizing higher-layer features of a deep network.
\newblock In {\em 2018 IEEE International Conference on Machine Learning
  Workshops}, 2009.

\bibitem{DBLP:conf/icml/EtmannLMS19}
Christian Etmann, Sebastian Lunz, Peter Maass, and Carola Sch{\"{o}}nlieb.
\newblock On the connection between adversarial robustness and saliency map
  interpretability.
\newblock In Kamalika Chaudhuri and Ruslan Salakhutdinov, editors, {\em
  Proceedings of the 36th International Conference on Machine Learning, {ICML}
  2019, 9-15 June 2019, Long Beach, California, {USA}}, volume~97 of {\em
  Proceedings of Machine Learning Research}, pages 1823--1832. {PMLR}, 2019.

\bibitem{DBLP:journals/corr/abs-2010-13663}
Lukas Faber, Amin~K. Moghaddam, and Roger Wattenhofer.
\newblock Contrastive graph neural network explanation.
\newblock {\em CoRR}, abs/2010.13663, 2020.

\bibitem{DBLP:conf/stoc/Feldman20}
Vitaly Feldman.
\newblock Does learning require memorization? a short tale about a long tail.
\newblock In Konstantin Makarychev, Yury Makarychev, Madhur Tulsiani, Gautam
  Kamath, and Julia Chuzhoy, editors, {\em Proccedings of the 52nd Annual {ACM}
  {SIGACT} Symposium on Theory of Computing, {STOC} 2020, Chicago, IL, USA,
  June 22-26, 2020}, pages 954--959. {ACM}, 2020.

\bibitem{DBLP:conf/nips/FeldmanZ20}
Vitaly Feldman and Chiyuan Zhang.
\newblock What neural networks memorize and why: Discovering the long tail via
  influence estimation.
\newblock In Hugo Larochelle, Marc'Aurelio Ranzato, Raia Hadsell,
  Maria{-}Florina Balcan, and Hsuan{-}Tien Lin, editors, {\em Advances in
  Neural Information Processing Systems 33: Annual Conference on Neural
  Information Processing Systems 2020, NeurIPS 2020, December 6-12, 2020,
  virtual}, 2020.

\bibitem{DBLP:conf/iccv/FongPV19}
Ruth Fong, Mandela Patrick, and Andrea Vedaldi.
\newblock Understanding deep networks via extremal perturbations and smooth
  masks.
\newblock In {\em 2019 {IEEE/CVF} International Conference on Computer Vision,
  {ICCV} 2019, Seoul, Korea (South), October 27 - November 2, 2019}, pages
  2950--2958. {IEEE}, 2019.

\bibitem{DBLP:conf/iccv/FongV17}
Ruth~C. Fong and Andrea Vedaldi.
\newblock Interpretable explanations of black boxes by meaningful perturbation.
\newblock In {\em {IEEE} International Conference on Computer Vision, {ICCV}
  2017, Venice, Italy, October 22-29, 2017}, pages 3449--3457. {IEEE} Computer
  Society, 2017.

\bibitem{DBLP:journals/corr/abs-1902-03501}
Sorelle~A. Friedler, Chitradeep~Dutta Roy, Carlos Scheidegger, and Dylan Slack.
\newblock Assessing the local interpretability of machine learning models.
\newblock {\em CoRR}, abs/1902.03501, 2019.

\bibitem{DBLP:conf/aiia/FrosstH17}
Nicholas Frosst and Geoffrey~E. Hinton.
\newblock Distilling a neural network into a soft decision tree.
\newblock In Tarek~R. Besold and Oliver Kutz, editors, {\em Proceedings of the
  First International Workshop on Comprehensibility and Explanation in {AI} and
  {ML} 2017 co-located with 16th International Conference of the Italian
  Association for Artificial Intelligence (AI*IA 2017), Bari, Italy, November
  16th and 17th, 2017}, volume 2071 of {\em {CEUR} Workshop Proceedings}.
  CEUR-WS.org, 2017.

\bibitem{DBLP:journals/natmi/GeirhosJMZBBW20}
Robert Geirhos, J{\"{o}}rn{-}Henrik Jacobsen, Claudio Michaelis, Richard~S.
  Zemel, Wieland Brendel, Matthias Bethge, and Felix~A. Wichmann.
\newblock Shortcut learning in deep neural networks.
\newblock {\em Nat. Mach. Intell.}, 2(11):665--673, 2020.

\bibitem{DBLP:conf/nips/GeirhosNMTBWB21}
Robert Geirhos, Kantharaju Narayanappa, Benjamin Mitzkus, Tizian Thieringer,
  Matthias Bethge, Felix~A. Wichmann, and Wieland Brendel.
\newblock Partial success in closing the gap between human and machine vision.
\newblock In Marc'Aurelio Ranzato, Alina Beygelzimer, Yann~N. Dauphin, Percy
  Liang, and Jennifer~Wortman Vaughan, editors, {\em Advances in Neural
  Information Processing Systems 34: Annual Conference on Neural Information
  Processing Systems 2021, NeurIPS 2021, December 6-14, 2021, virtual}, pages
  23885--23899, 2021.

\bibitem{DBLP:conf/iclr/GeirhosRMBWB19}
Robert Geirhos, Patricia Rubisch, Claudio Michaelis, Matthias Bethge, Felix~A.
  Wichmann, and Wieland Brendel.
\newblock Imagenet-trained cnns are biased towards texture; increasing shape
  bias improves accuracy and robustness.
\newblock In {\em 7th International Conference on Learning Representations,
  {ICLR} 2019, New Orleans, LA, USA, May 6-9, 2019}. OpenReview.net, 2019.

\bibitem{DBLP:conf/emnlp/GhaeiniFT18}
Reza Ghaeini, Xiaoli~Z. Fern, and Prasad Tadepalli.
\newblock Interpreting recurrent and attention-based neural models: a case
  study on natural language inference.
\newblock In Ellen Riloff, David Chiang, Julia Hockenmaier, and Jun'ichi
  Tsujii, editors, {\em Proceedings of the 2018 Conference on Empirical Methods
  in Natural Language Processing}. Association for Computational Linguistics,
  2018.

\bibitem{DBLP:journals/jcc/GohHV17}
Garrett~B. Goh, Nathan~O. Hodas, and Abhinav Vishnu.
\newblock Deep learning for computational chemistry.
\newblock {\em J. Comput. Chem.}, 38(16):1291--1307, 2017.

\bibitem{DBLP:journals/tmis/Gomez-UribeH16}
Carlos~Alberto Gomez{-}Uribe and Neil Hunt.
\newblock The netflix recommender system: Algorithms, business value, and
  innovation.
\newblock {\em {ACM} Trans. Manag. Inf. Syst.}, 6(4):13:1--13:19, 2016.

\bibitem{DBLP:conf/icml/GoyalWEBPL19}
Yash Goyal, Ziyan Wu, Jan Ernst, Dhruv Batra, Devi Parikh, and Stefan Lee.
\newblock Counterfactual visual explanations.
\newblock In Kamalika Chaudhuri and Ruslan Salakhutdinov, editors, {\em
  Proceedings of the 36th International Conference on Machine Learning, {ICML}
  2019, 9-15 June 2019, Long Beach, California, {USA}}, volume~97 of {\em
  Proceedings of Machine Learning Research}, pages 2376--2384. {PMLR}, 2019.

\bibitem{DBLP:conf/icml/GreydanusKDF18}
Samuel Greydanus, Anurag Koul, Jonathan Dodge, and Alan Fern.
\newblock Visualizing and understanding atari agents.
\newblock In Jennifer~G. Dy and Andreas Krause, editors, {\em Proceedings of
  the 35th International Conference on Machine Learning, {ICML} 2018,
  Stockholmsm{\"{a}}ssan, Stockholm, Sweden, July 10-15, 2018}, volume~80 of
  {\em Proceedings of Machine Learning Research}, pages 1787--1796. {PMLR},
  2018.

\bibitem{DBLP:conf/www/Grgic-HlacaRGW18}
Nina Grgic{-}Hlaca, Elissa~M. Redmiles, Krishna~P. Gummadi, and Adrian Weller.
\newblock Human perceptions of fairness in algorithmic decision making: {A}
  case study of criminal risk prediction.
\newblock In Pierre{-}Antoine Champin, Fabien Gandon, Mounia Lalmas, and
  Panagiotis~G. Ipeirotis, editors, {\em Proceedings of the 2018 World Wide Web
  Conference on World Wide Web, {WWW} 2018, Lyon, France, April 23-27, 2018},
  pages 903--912. {ACM}, 2018.

\bibitem{DBLP:conf/accv/GuYT18}
Jindong Gu, Yinchong Yang, and Volker Tresp.
\newblock Understanding individual decisions of cnns via contrastive
  backpropagation.
\newblock In C.~V. Jawahar, Hongdong Li, Greg Mori, and Konrad Schindler,
  editors, {\em Computer Vision - {ACCV} 2018 - 14th Asian Conference on
  Computer Vision, Perth, Australia, December 2-6, 2018, Revised Selected
  Papers, Part {III}}, volume 11363 of {\em Lecture Notes in Computer Science},
  pages 119--134. Springer, 2018.

\bibitem{DBLP:journals/corr/abs-1708-06733}
Tianyu Gu, Brendan Dolan{-}Gavitt, and Siddharth Garg.
\newblock {BadNets}: Identifying vulnerabilities in the machine learning model
  supply chain.
\newblock {\em CoRR}, abs/1708.06733, 2017.

\bibitem{DBLP:conf/pkdd/GuidottiMMP19}
Riccardo Guidotti, Anna Monreale, Stan Matwin, and Dino Pedreschi.
\newblock Black box explanation by learning image exemplars in the latent
  feature space.
\newblock In Ulf Brefeld, {\'{E}}lisa Fromont, Andreas Hotho, Arno~J. Knobbe,
  Marloes~H. Maathuis, and C{\'{e}}line Robardet, editors, {\em Machine
  Learning and Knowledge Discovery in Databases - European Conference, {ECML}
  {PKDD} 2019, W{\"{u}}rzburg, Germany, September 16-20, 2019, Proceedings,
  Part {I}}, volume 11906 of {\em Lecture Notes in Computer Science}, pages
  189--205. Springer, 2019.

\bibitem{DBLP:conf/iclr/HendrycksD19}
Dan Hendrycks and Thomas~G. Dietterich.
\newblock Benchmarking neural network robustness to common corruptions and
  perturbations.
\newblock In {\em 7th International Conference on Learning Representations,
  {ICLR} 2019, New Orleans, LA, USA, May 6-9, 2019}. OpenReview.net, 2019.

\bibitem{DBLP:conf/nips/HeoJM19}
Juyeon Heo, Sunghwan Joo, and Taesup Moon.
\newblock Fooling neural network interpretations via adversarial model
  manipulation.
\newblock In Hanna~M. Wallach, Hugo Larochelle, Alina Beygelzimer, Florence
  d'Alch{\'{e}}{-}Buc, Emily~B. Fox, and Roman Garnett, editors, {\em Advances
  in Neural Information Processing Systems 32: Annual Conference on Neural
  Information Processing Systems 2019, NeurIPS 2019, December 8-14, 2019,
  Vancouver, BC, Canada}, pages 2921--2932, 2019.

\bibitem{DBLP:conf/cscw/HerlockerKR00}
Jonathan~L. Herlocker, Joseph~A. Konstan, and John Riedl.
\newblock Explaining collaborative filtering recommendations.
\newblock In Wendy~A. Kellogg and Steve Whittaker, editors, {\em {CSCW} 2000,
  Proceeding on the {ACM} 2000 Conference on Computer Supported Cooperative
  Work, Philadelphia, PA, USA, December 2-6, 2000}, pages 241--250. {ACM},
  2000.

\bibitem{DBLP:conf/iclr/HintonSF18}
Geoffrey~E. Hinton, Sara Sabour, and Nicholas Frosst.
\newblock Matrix capsules with {EM} routing.
\newblock In {\em 6th International Conference on Learning Representations,
  {ICLR} 2018, Vancouver, BC, Canada, April 30 - May 3, 2018, Conference Track
  Proceedings}. OpenReview.net, 2018.

\bibitem{DBLP:conf/nips/HookerEKK19}
Sara Hooker, Dumitru Erhan, Pieter{-}Jan Kindermans, and Been Kim.
\newblock A benchmark for interpretability methods in deep neural networks.
\newblock In Hanna~M. Wallach, Hugo Larochelle, Alina Beygelzimer, Florence
  d'Alch{\'{e}}{-}Buc, Emily~B. Fox, and Roman Garnett, editors, {\em Advances
  in Neural Information Processing Systems 32: Annual Conference on Neural
  Information Processing Systems 2019, NeurIPS 2019, December 8-14, 2019,
  Vancouver, BC, Canada}, pages 9734--9745, 2019.

\bibitem{hua2015computer}
Kai-Lung Hua, Che-Hao Hsu, Shintami~Chusnul Hidayati, Wen-Huang Cheng, and
  Yu-Jen Chen.
\newblock Computer-aided classification of lung nodules on computed tomography
  images via deep learning technique.
\newblock {\em OncoTargets and therapy}, 2015.

\bibitem{DBLP:journals/corr/abs-2001-06216}
Qiang Huang, Makoto Yamada, Yuan Tian, Dinesh Singh, Dawei Yin, and Yi~Chang.
\newblock Graphlime: Local interpretable model explanations for graph neural
  networks.
\newblock {\em CoRR}, abs/2001.06216, 2020.

\bibitem{DBLP:conf/nips/IlyasSTETM19}
Andrew Ilyas, Shibani Santurkar, Dimitris Tsipras, Logan Engstrom, Brandon
  Tran, and Aleksander Madry.
\newblock Adversarial examples are not bugs, they are features.
\newblock In Hanna~M. Wallach, Hugo Larochelle, Alina Beygelzimer, Florence
  d'Alch{\'{e}}{-}Buc, Emily~B. Fox, and Roman Garnett, editors, {\em Advances
  in Neural Information Processing Systems 32: Annual Conference on Neural
  Information Processing Systems 2019, NeurIPS 2019, December 8-14, 2019,
  Vancouver, BC, Canada}, pages 125--136, 2019.

\bibitem{DBLP:conf/flairs/IslamEG20}
Sheikh~Rabiul Islam, William Eberle, and Sheikh~K. Ghafoor.
\newblock Towards quantification of explainability in explainable artificial
  intelligence methods.
\newblock In Roman Bart{\'{a}}k and Eric Bell, editors, {\em Proceedings of the
  Thirty-Third International Florida Artificial Intelligence Research Society
  Conference, Originally to be held in North Miami Beach, Florida, USA, May
  17-20, 2020}, pages 75--81. {AAAI} Press, 2020.

\bibitem{DBLP:conf/iccvw/IwanaKU19}
Brian~Kenji Iwana, Ryohei Kuroki, and Seiichi Uchida.
\newblock Explaining convolutional neural networks using softmax gradient
  layer-wise relevance propagation.
\newblock In {\em 2019 {IEEE/CVF} International Conference on Computer Vision
  Workshops, {ICCV} Workshops 2019, Seoul, Korea (South), October 27-28, 2019},
  pages 4176--4185. {IEEE}, 2019.

\bibitem{DBLP:conf/aies/IyerLL0SS18}
Rahul Iyer, Yuezhang Li, Huao Li, Michael Lewis, Ramitha Sundar, and Katia~P.
  Sycara.
\newblock Transparency and explanation in deep reinforcement learning neural
  networks.
\newblock In Jason Furman, Gary~E. Marchant, Huw Price, and Francesca Rossi,
  editors, {\em Proceedings of the 2018 {AAAI/ACM} Conference on AI, Ethics,
  and Society, {AIES} 2018, New Orleans, LA, USA, February 02-03, 2018}, pages
  144--150. {ACM}, 2018.

\bibitem{DBLP:conf/acl/JacoviG20}
Alon Jacovi and Yoav Goldberg.
\newblock Towards faithfully interpretable {NLP} systems: How should we define
  and evaluate faithfulness?
\newblock In Dan Jurafsky, Joyce Chai, Natalie Schluter, and Joel~R. Tetreault,
  editors, {\em Proceedings of the 58th Annual Meeting of the Association for
  Computational Linguistics, {ACL} 2020, Online, July 5-10, 2020}, pages
  4198--4205. Association for Computational Linguistics, 2020.

\bibitem{DBLP:conf/naacl/JainW19}
Sarthak Jain and Byron~C. Wallace.
\newblock Attention is not explanation.
\newblock In Jill Burstein, Christy Doran, and Thamar Solorio, editors, {\em
  Proceedings of the 2019 Conference of the North American Chapter of the
  Association for Computational Linguistics: Human Language Technologies,
  {NAACL-HLT}}. Association for Computational Linguistics, 2019.

\bibitem{DBLP:journals/corr/abs-1905-00931}
Taeho Jo, Kwangsik Nho, and Andrew~J. Saykin.
\newblock Deep learning in {Alzheimer}'s disease: Diagnostic classification and
  prognostic prediction using neuroimaging data.
\newblock {\em CoRR}, abs/1905.00931, 2019.

\bibitem{jumper2021highly}
John Jumper, Richard Evans, Alexander Pritzel, Tim Green, Michael Figurnov,
  Olaf Ronneberger, Kathryn Tunyasuvunakool, Russ Bates, Augustin
  {\v{Z}}{\'\i}dek, Anna Potapenko, et~al.
\newblock Highly accurate protein structure prediction with alphafold.
\newblock {\em Nature}, 2021.

\bibitem{DBLP:journals/jair/KaelblingLM96}
Leslie~Pack Kaelbling, Michael~L. Littman, and Andrew~W. Moore.
\newblock Reinforcement learning: {A} survey.
\newblock {\em J. Artif. Intell. Res.}, 4:237--285, 1996.

\bibitem{DBLP:journals/corr/abs-2110-06968}
Asad Khan, E.~A. Huerta, and Huihuo Zheng.
\newblock Interpretable {AI} forecasting for numerical relativity waveforms of
  quasi-circular, spinning, non-precessing binary black hole mergers.
\newblock {\em CoRR}, abs/2110.06968, 2021.

\bibitem{DBLP:conf/icml/KimWGCWVS18}
Been Kim, Martin Wattenberg, Justin Gilmer, Carrie~J. Cai, James Wexler,
  Fernanda~B. Vi{\'{e}}gas, and Rory Sayres.
\newblock Interpretability beyond feature attribution: Quantitative testing
  with concept activation vectors {(TCAV)}.
\newblock In Jennifer~G. Dy and Andreas Krause, editors, {\em Proceedings of
  the 35th International Conference on Machine Learning, {ICML} 2018,
  Stockholmsm{\"{a}}ssan, Stockholm, Sweden, July 10-15, 2018}, volume~80 of
  {\em Proceedings of Machine Learning Research}, pages 2673--2682. {PMLR},
  2018.

\bibitem{DBLP:conf/iccv/KimCAO21}
Jae{-}Myung Kim, Junsuk Choe, Zeynep Akata, and Seong~Joon Oh.
\newblock Keep {CALM} and improve visual feature attribution.
\newblock In {\em 2021 {IEEE/CVF} International Conference on Computer Vision,
  {ICCV} 2021, Montreal, QC, Canada, October 10-17, 2021}, pages 8330--8340.
  {IEEE}, 2021.

\bibitem{DBLP:conf/icml/KimCS20}
Jang{-}Hyun Kim, Wonho Choo, and Hyun~Oh Song.
\newblock Puzzle mix: Exploiting saliency and local statistics for optimal
  mixup.
\newblock In {\em Proceedings of the 37th International Conference on Machine
  Learning, {ICML} 2020, 13-18 July 2020, Virtual Event}, volume 119 of {\em
  Proceedings of Machine Learning Research}, pages 5275--5285. {PMLR}, 2020.

\bibitem{DBLP:conf/nips/KohATL19}
Pang~Wei Koh, Kai{-}Siang Ang, Hubert H.~K. Teo, and Percy Liang.
\newblock On the accuracy of influence functions for measuring group effects.
\newblock In Hanna~M. Wallach, Hugo Larochelle, Alina Beygelzimer, Florence
  d'Alch{\'{e}}{-}Buc, Emily~B. Fox, and Roman Garnett, editors, {\em Advances
  in Neural Information Processing Systems 32: Annual Conference on Neural
  Information Processing Systems 2019, NeurIPS 2019, December 8-14, 2019,
  Vancouver, BC, Canada}, pages 5255--5265, 2019.

\bibitem{DBLP:conf/icml/KohL17}
Pang~Wei Koh and Percy Liang.
\newblock Understanding black-box predictions via influence functions.
\newblock In Doina Precup and Yee~Whye Teh, editors, {\em Proceedings of the
  34th International Conference on Machine Learning, {ICML} 2017, Sydney, NSW,
  Australia, 6-11 August 2017}, volume~70 of {\em Proceedings of Machine
  Learning Research}, pages 1885--1894. {PMLR}, 2017.

\bibitem{DBLP:conf/iccv/KontschiederFCB15}
Peter Kontschieder, Madalina Fiterau, Antonio Criminisi, and Samuel~Rota
  Bul{\`{o}}.
\newblock Deep neural decision forests.
\newblock In {\em 2015 {IEEE} International Conference on Computer Vision,
  {ICCV} 2015, Santiago, Chile, December 7-13, 2015}, pages 1467--1475. {IEEE}
  Computer Society, 2015.

\bibitem{DBLP:conf/acl/KumarT20}
Sawan Kumar and Partha~P. Talukdar.
\newblock {NILE} : Natural language inference with faithful natural language
  explanations.
\newblock In Dan Jurafsky, Joyce Chai, Natalie Schluter, and Joel~R. Tetreault,
  editors, {\em Proceedings of the 58th Annual Meeting of the Association for
  Computational Linguistics, {ACL}}. Association for Computational Linguistics,
  2020.

\bibitem{DBLP:journals/corr/abs-1902-00006}
Isaac Lage, Emily Chen, Jeffrey He, Menaka Narayanan, Been Kim, Sam Gershman,
  and Finale Doshi{-}Velez.
\newblock An evaluation of the human-interpretability of explanation.
\newblock {\em CoRR}, abs/1902.00006, 2019.

\bibitem{DBLP:conf/ijcai/LaiG17}
Baisheng Lai and Xiaojin Gong.
\newblock Saliency guided end-to-end learning for weakly supervised object
  detection.
\newblock In Carles Sierra, editor, {\em Proceedings of the Twenty-Sixth
  International Joint Conference on Artificial Intelligence, {IJCAI} 2017,
  Melbourne, Australia, August 19-25, 2017}, pages 2053--2059. ijcai.org, 2017.

\bibitem{DBLP:journals/corr/LakkarajuKCL17}
Himabindu Lakkaraju, Ece Kamar, Rich Caruana, and Jure Leskovec.
\newblock Interpretable {\&} explorable approximations of black box models.
\newblock {\em CoRR}, abs/1707.01154, 2017.

\bibitem{DBLP:conf/pkdd/LaugelLMRD19}
Thibault Laugel, Marie{-}Jeanne Lesot, Christophe Marsala, Xavier Renard, and
  Marcin Detyniecki.
\newblock Unjustified classification regions and counterfactual explanations in
  machine learning.
\newblock In Ulf Brefeld, {\'{E}}lisa Fromont, Andreas Hotho, Arno~J. Knobbe,
  Marloes~H. Maathuis, and C{\'{e}}line Robardet, editors, {\em Machine
  Learning and Knowledge Discovery in Databases - European Conference, {ECML}
  {PKDD} 2019, W{\"{u}}rzburg, Germany, September 16-20, 2019, Proceedings,
  Part {II}}, volume 11907 of {\em Lecture Notes in Computer Science}, pages
  37--54. Springer, 2019.

\bibitem{DBLP:journals/nature/LeCunBH15}
Yann LeCun, Yoshua Bengio, and Geoffrey~E. Hinton.
\newblock Deep learning.
\newblock {\em Nat.}, 521(7553):436--444, 2015.

\bibitem{DBLP:journals/jmlr/LevineFDA16}
Sergey Levine, Chelsea Finn, Trevor Darrell, and Pieter Abbeel.
\newblock End-to-end training of deep visuomotor policies.
\newblock {\em J. Mach. Learn. Res.}, 17:39:1--39:40, 2016.

\bibitem{DBLP:journals/ijrr/LevinePKIQ18}
Sergey Levine, Peter Pastor, Alex Krizhevsky, Julian Ibarz, and Deirdre
  Quillen.
\newblock Learning hand-eye coordination for robotic grasping with deep
  learning and large-scale data collection.
\newblock {\em Int. J. Robotics Res.}, 37(4-5):421--436, 2018.

\bibitem{DBLP:journals/corr/abs-2110-01167}
Bo~Li, Peng Qi, Bo~Liu, Shuai Di, Jingen Liu, Jiquan Pei, Jinfeng Yi, and Bowen
  Zhou.
\newblock Trustworthy {AI:} from principles to practices.
\newblock {\em CoRR}, abs/2110.01167, 2021.

\bibitem{DBLP:conf/sigir/LiQPQDW19}
Chenliang Li, Cong Quan, Li~Peng, Yunwei Qi, Yuming Deng, and Libing Wu.
\newblock A capsule network for recommendation and explaining what you like and
  dislike.
\newblock In Benjamin Piwowarski, Max Chevalier, {\'{E}}ric Gaussier, Yoelle
  Maarek, Jian{-}Yun Nie, and Falk Scholer, editors, {\em Proceedings of the
  42nd International {ACM} {SIGIR} Conference on Research and Development in
  Information Retrieval, {SIGIR} 2019, Paris, France, July 21-25, 2019}, pages
  275--284. {ACM}, 2019.

\bibitem{DBLP:conf/aaai/LiLCR18}
Oscar Li, Hao Liu, Chaofan Chen, and Cynthia Rudin.
\newblock Deep learning for case-based reasoning through prototypes: {A} neural
  network that explains its predictions.
\newblock In Sheila~A. McIlraith and Kilian~Q. Weinberger, editors, {\em
  Proceedings of the Thirty-Second {AAAI} Conference on Artificial
  Intelligence, (AAAI-18), the 30th innovative Applications of Artificial
  Intelligence (IAAI-18), and the 8th {AAAI} Symposium on Educational Advances
  in Artificial Intelligence (EAAI-18), New Orleans, Louisiana, USA, February
  2-7, 2018}, pages 3530--3537. {AAAI} Press, 2018.

\bibitem{DBLP:journals/corr/abs-2109-00707}
Xuhong Li, Haoyi Xiong, Siyu Huang, Shilei Ji, and Dejing Dou.
\newblock Cross-model consensus of explanations and beyond for image
  classification models: An empirical study.
\newblock {\em CoRR}, abs/2109.00707, 2021.

\bibitem{DBLP:journals/corr/Li17b}
Yuxi Li.
\newblock Deep reinforcement learning: An overview.
\newblock {\em CoRR}, abs/1701.07274, 2017.

\bibitem{DBLP:conf/kdd/LinLC21}
Yi{-}Shan Lin, Wen{-}Chuan Lee, and Z.~Berkay Celik.
\newblock What do you see?: Evaluation of explainable artificial intelligence
  {(XAI)} interpretability through neural backdoors.
\newblock In Feida Zhu, Beng~Chin Ooi, and Chunyan Miao, editors, {\em {KDD}
  '21: The 27th {ACM} {SIGKDD} Conference on Knowledge Discovery and Data
  Mining, Virtual Event, Singapore, August 14-18, 2021}, pages 1027--1035.
  {ACM}, 2021.

\bibitem{DBLP:journals/cacm/Lipton18}
Zachary~C. Lipton.
\newblock The mythos of model interpretability.
\newblock {\em Commun. {ACM}}, 61(10):36--43, 2018.

\bibitem{DBLP:journals/mia/LitjensKBSCGLGS17}
Geert Litjens, Thijs Kooi, Babak~Ehteshami Bejnordi, Arnaud Arindra~Adiyoso
  Setio, Francesco Ciompi, Mohsen Ghafoorian, Jeroen A. W.~M. van~der Laak,
  Bram van Ginneken, and Clara~I. S{\'{a}}nchez.
\newblock A survey on deep learning in medical image analysis.
\newblock {\em Medical Image Anal.}, 42:60--88, 2017.

\bibitem{DBLP:conf/acl/LiuYW19}
Hui Liu, Qingyu Yin, and William~Yang Wang.
\newblock Towards explainable {NLP:} {A} generative explanation framework for
  text classification.
\newblock In Anna Korhonen, David~R. Traum, and Llu{\'{\i}}s M{\`{a}}rquez,
  editors, {\em Proceedings of the 57th Conference of the Association for
  Computational Linguistics, {ACL}}. Association for Computational Linguistics,
  2019.

\bibitem{DBLP:conf/nips/LundbergL17}
Scott~M. Lundberg and Su{-}In Lee.
\newblock A unified approach to interpreting model predictions.
\newblock In Isabelle Guyon, Ulrike von Luxburg, Samy Bengio, Hanna~M. Wallach,
  Rob Fergus, S.~V.~N. Vishwanathan, and Roman Garnett, editors, {\em Advances
  in Neural Information Processing Systems 30: Annual Conference on Neural
  Information Processing Systems 2017, December 4-9, 2017, Long Beach, CA,
  {USA}}, pages 4765--4774, 2017.

\bibitem{DBLP:conf/nips/LuoCXYZC020}
Dongsheng Luo, Wei Cheng, Dongkuan Xu, Wenchao Yu, Bo~Zong, Haifeng Chen, and
  Xiang Zhang.
\newblock Parameterized explainer for graph neural network.
\newblock In Hugo Larochelle, Marc'Aurelio Ranzato, Raia Hadsell,
  Maria{-}Florina Balcan, and Hsuan{-}Tien Lin, editors, {\em Advances in
  Neural Information Processing Systems 33: Annual Conference on Neural
  Information Processing Systems 2020, NeurIPS 2020, December 6-12, 2020,
  virtual}, 2020.

\bibitem{ma2019paddlepaddle}
Yanjun Ma, Dianhai Yu, Tian Wu, and Haifeng Wang.
\newblock Paddlepaddle: An open-source deep learning platform from industrial
  practice.
\newblock {\em Frontiers of Data and Domputing}, 2019.

\bibitem{DBLP:conf/cvpr/MahendranV15}
Aravindh Mahendran and Andrea Vedaldi.
\newblock Understanding deep image representations by inverting them.
\newblock In {\em {IEEE} Conference on Computer Vision and Pattern Recognition,
  {CVPR} 2015, Boston, MA, USA, June 7-12, 2015}, pages 5188--5196. {IEEE}
  Computer Society, 2015.

\bibitem{DBLP:journals/corr/abs-2012-01166}
Andrei Margeloiu, Nikola Simidjievski, Mateja Jamnik, and Adrian Weller.
\newblock Improving interpretability in medical imaging diagnosis using
  adversarial training.
\newblock {\em CoRR}, abs/2012.01166, 2020.

\bibitem{DBLP:journals/ai/Miller19}
Tim Miller.
\newblock Explanation in artificial intelligence: Insights from the social
  sciences.
\newblock {\em Artif. Intell.}, 267:1--38, 2019.

\bibitem{DBLP:conf/kdd/MingXQR19}
Yao Ming, Panpan Xu, Huamin Qu, and Liu Ren.
\newblock Interpretable and steerable sequence learning via prototypes.
\newblock In Ankur Teredesai, Vipin Kumar, Ying Li, R{\'{o}}mer Rosales,
  Evimaria Terzi, and George Karypis, editors, {\em Proceedings of the 25th
  {ACM} {SIGKDD} International Conference on Knowledge Discovery {\&} Data
  Mining, {KDD} 2019, Anchorage, AK, USA, August 4-8, 2019}, pages 903--913.
  {ACM}, 2019.

\bibitem{DBLP:journals/nature/MnihKSRVBGRFOPB15}
Volodymyr Mnih, Koray Kavukcuoglu, David Silver, Andrei~A. Rusu, Joel Veness,
  Marc~G. Bellemare, Alex Graves, Martin~A. Riedmiller, Andreas Fidjeland,
  Georg Ostrovski, Stig Petersen, Charles Beattie, Amir Sadik, Ioannis
  Antonoglou, Helen King, Dharshan Kumaran, Daan Wierstra, Shane Legg, and
  Demis Hassabis.
\newblock Human-level control through deep reinforcement learning.
\newblock {\em Nat.}, 518(7540):529--533, 2015.

\bibitem{DBLP:journals/pr/MontavonLBSM17}
Gr{\'{e}}goire Montavon, Sebastian Lapuschkin, Alexander Binder, Wojciech
  Samek, and Klaus{-}Robert M{\"{u}}ller.
\newblock Explaining nonlinear classification decisions with deep taylor
  decomposition.
\newblock {\em Pattern Recognit.}, 65:211--222, 2017.

\bibitem{DBLP:journals/dsp/MontavonSM18}
Gr{\'{e}}goire Montavon, Wojciech Samek, and Klaus{-}Robert M{\"{u}}ller.
\newblock Methods for interpreting and understanding deep neural networks.
\newblock {\em Digit. Signal Process.}, 73:1--15, 2018.

\bibitem{DBLP:journals/sigkdd/MoraffahKGRL20}
Raha Moraffah, Mansooreh Karami, Ruocheng Guo, Adrienne Raglin, and Huan Liu.
\newblock Causal interpretability for machine learning - problems, methods and
  evaluation.
\newblock {\em {SIGKDD} Explor.}, 22(1):18--33, 2020.

\bibitem{DBLP:conf/fat/MothilalST20}
Ramaravind~Kommiya Mothilal, Amit Sharma, and Chenhao Tan.
\newblock Explaining machine learning classifiers through diverse
  counterfactual explanations.
\newblock In Mireille Hildebrandt, Carlos Castillo, L.~Elisa Celis, Salvatore
  Ruggieri, Linnet Taylor, and Gabriela Zanfir{-}Fortuna, editors, {\em FAT*
  '20: Conference on Fairness, Accountability, and Transparency, Barcelona,
  Spain, January 27-30, 2020}, pages 607--617. {ACM}, 2020.

\bibitem{DBLP:journals/corr/abs-1901-04592}
W.~James Murdoch, Chandan Singh, Karl Kumbier, Reza Abbasi{-}Asl, and Bin Yu.
\newblock Interpretable machine learning: definitions, methods, and
  applications.
\newblock {\em CoRR}, abs/1901.04592, 2019.

\bibitem{DBLP:conf/aaai/NamGCWL20}
Woo{-}Jeoung Nam, Shir Gur, Jaesik Choi, Lior Wolf, and Seong{-}Whan Lee.
\newblock Relative attributing propagation: Interpreting the comparative
  contributions of individual units in deep neural networks.
\newblock In {\em The Thirty-Fourth {AAAI} Conference on Artificial
  Intelligence, {AAAI} 2020, The Thirty-Second Innovative Applications of
  Artificial Intelligence Conference, {IAAI} 2020, The Tenth {AAAI} Symposium
  on Educational Advances in Artificial Intelligence, {EAAI} 2020, New York,
  NY, USA, February 7-12, 2020}, pages 2501--2508. {AAAI} Press, 2020.

\bibitem{DBLP:conf/nips/NguyenDYBC16}
Anh~Mai Nguyen, Alexey Dosovitskiy, Jason Yosinski, Thomas Brox, and Jeff
  Clune.
\newblock Synthesizing the preferred inputs for neurons in neural networks via
  deep generator networks.
\newblock In Daniel~D. Lee, Masashi Sugiyama, Ulrike von Luxburg, Isabelle
  Guyon, and Roman Garnett, editors, {\em Advances in Neural Information
  Processing Systems 29: Annual Conference on Neural Information Processing
  Systems 2016, December 5-10, 2016, Barcelona, Spain}, pages 3387--3395, 2016.

\bibitem{DBLP:conf/nips/PaszkeGMLBCKLGA19}
Adam Paszke, Sam Gross, Francisco Massa, Adam Lerer, James Bradbury, Gregory
  Chanan, Trevor Killeen, Zeming Lin, Natalia Gimelshein, Luca Antiga, Alban
  Desmaison, Andreas K{\"{o}}pf, Edward~Z. Yang, Zachary DeVito, Martin Raison,
  Alykhan Tejani, Sasank Chilamkurthy, Benoit Steiner, Lu~Fang, Junjie Bai, and
  Soumith Chintala.
\newblock Pytorch: An imperative style, high-performance deep learning library.
\newblock In Hanna~M. Wallach, Hugo Larochelle, Alina Beygelzimer, Florence
  d'Alch{\'{e}}{-}Buc, Emily~B. Fox, and Roman Garnett, editors, {\em Advances
  in Neural Information Processing Systems 32: Annual Conference on Neural
  Information Processing Systems 2019, NeurIPS 2019, December 8-14, 2019,
  Vancouver, BC, Canada}, pages 8024--8035, 2019.

\bibitem{pearl2009causal}
Judea Pearl et~al.
\newblock Causal inference in statistics: An overview.
\newblock {\em Statistics surveys}, 2009.

\bibitem{DBLP:conf/bmvc/PetsiukDS18}
Vitali Petsiuk, Abir Das, and Kate Saenko.
\newblock {RISE:} randomized input sampling for explanation of black-box
  models.
\newblock In {\em British Machine Vision Conference 2018, {BMVC} 2018,
  Newcastle, UK, September 3-6, 2018}, page 151. {BMVA} Press, 2018.

\bibitem{DBLP:conf/nips/Pleiss0EW20}
Geoff Pleiss, Tianyi Zhang, Ethan~R. Elenberg, and Kilian~Q. Weinberger.
\newblock Identifying mislabeled data using the area under the margin ranking.
\newblock In Hugo Larochelle, Marc'Aurelio Ranzato, Raia Hadsell,
  Maria{-}Florina Balcan, and Hsuan{-}Tien Lin, editors, {\em Advances in
  Neural Information Processing Systems 33: Annual Conference on Neural
  Information Processing Systems 2020, NeurIPS 2020, December 6-12, 2020,
  virtual}, 2020.

\bibitem{DBLP:conf/nips/PlumbACPXT20}
Gregory Plumb, Maruan Al{-}Shedivat, {\'{A}}ngel~Alexander Cabrera, Adam Perer,
  Eric~P. Xing, and Ameet Talwalkar.
\newblock Regularizing black-box models for improved interpretability.
\newblock In Hugo Larochelle, Marc'Aurelio Ranzato, Raia Hadsell,
  Maria{-}Florina Balcan, and Hsuan{-}Tien Lin, editors, {\em Advances in
  Neural Information Processing Systems 33: Annual Conference on Neural
  Information Processing Systems 2020, NeurIPS 2020, December 6-12, 2020,
  virtual}, 2020.

\bibitem{DBLP:conf/nips/PlumbMT18}
Gregory Plumb, Denali Molitor, and Ameet Talwalkar.
\newblock Model agnostic supervised local explanations.
\newblock In Samy Bengio, Hanna~M. Wallach, Hugo Larochelle, Kristen Grauman,
  Nicol{\`{o}} Cesa{-}Bianchi, and Roman Garnett, editors, {\em Advances in
  Neural Information Processing Systems 31: Annual Conference on Neural
  Information Processing Systems 2018, NeurIPS 2018, December 3-8, 2018,
  Montr{\'{e}}al, Canada}, pages 2520--2529, 2018.

\bibitem{DBLP:conf/iclr/PlumeraultBH20}
Antoine Plumerault, Herv{\'{e}}~Le Borgne, and C{\'{e}}line Hudelot.
\newblock Controlling generative models with continuous factors of variations.
\newblock In {\em 8th International Conference on Learning Representations,
  {ICLR} 2020, Addis Ababa, Ethiopia, April 26-30, 2020}. OpenReview.net, 2020.

\bibitem{DBLP:conf/cvpr/PopeKRMH19}
Phillip~E. Pope, Soheil Kolouri, Mohammad Rostami, Charles~E. Martin, and Heiko
  Hoffmann.
\newblock Explainability methods for graph convolutional neural networks.
\newblock In {\em {IEEE} Conference on Computer Vision and Pattern Recognition,
  {CVPR} 2019, Long Beach, CA, USA, June 16-20, 2019}, pages 10772--10781.
  Computer Vision Foundation / {IEEE}, 2019.

\bibitem{DBLP:series/lncs/PreuerKRHU19}
Kristina Preuer, G{\"{u}}nter Klambauer, Friedrich Rippmann, Sepp Hochreiter,
  and Thomas Unterthiner.
\newblock Interpretable deep learning in drug discovery.
\newblock In Wojciech Samek, Gr{\'{e}}goire Montavon, Andrea Vedaldi, Lars~Kai
  Hansen, and Klaus{-}Robert M{\"{u}}ller, editors, {\em Explainable {AI:}
  Interpreting, Explaining and Visualizing Deep Learning}, volume 11700 of {\em
  Lecture Notes in Computer Science}, pages 331--345. Springer, 2019.

\bibitem{DBLP:conf/cdmake/PuiuttaV20}
Erika Puiutta and Eric M. S.~P. Veith.
\newblock Explainable reinforcement learning: {A} survey.
\newblock In Andreas Holzinger, Peter Kieseberg, A~Min Tjoa, and Edgar~R.
  Weippl, editors, {\em Machine Learning and Knowledge Extraction - 4th {IFIP}
  {TC} 5, {TC} 12, {WG} 8.4, {WG} 8.9, {WG} 12.9 International Cross-Domain
  Conference, {CD-MAKE} 2020, Dublin, Ireland, August 25-28, 2020,
  Proceedings}, volume 12279 of {\em Lecture Notes in Computer Science}, pages
  77--95. Springer, 2020.

\bibitem{DBLP:conf/iclr/PuriVGKDK020}
Nikaash Puri, Sukriti Verma, Piyush Gupta, Dhruv Kayastha, Shripad Deshmukh,
  Balaji Krishnamurthy, and Sameer Singh.
\newblock Explain your move: Understanding agent actions using specific and
  relevant feature attribution.
\newblock In {\em 8th International Conference on Learning Representations,
  {ICLR} 2020, Addis Ababa, Ethiopia, April 26-30, 2020}. OpenReview.net, 2020.

\bibitem{rajpurkar2020chexaid}
Pranav Rajpurkar, Chloe O’Connell, Amit Schechter, Nishit Asnani, Jason Li,
  Amirhossein Kiani, Robyn~L Ball, Marc Mendelson, Gary Maartens, Dani{\"e}l~J
  van Hoving, et~al.
\newblock Chexaid: deep learning assistance for physician diagnosis of
  tuberculosis using chest x-rays in patients with {HIV}.
\newblock {\em {NPJ} digital medicine}, 2020.

\bibitem{DBLP:conf/kdd/Ribeiro0G16}
Marco~T{\'{u}}lio Ribeiro, Sameer Singh, and Carlos Guestrin.
\newblock ``{Why} should {I} trust you?'': Explaining the predictions of any
  classifier.
\newblock In Balaji Krishnapuram, Mohak Shah, Alexander~J. Smola, Charu~C.
  Aggarwal, Dou Shen, and Rajeev Rastogi, editors, {\em Proceedings of the 22nd
  {ACM} {SIGKDD} International Conference on Knowledge Discovery and Data
  Mining, San Francisco, CA, USA, August 13-17, 2016}, pages 1135--1144. {ACM},
  2016.

\bibitem{DBLP:conf/aaai/Ribeiro0G18}
Marco~T{\'{u}}lio Ribeiro, Sameer Singh, and Carlos Guestrin.
\newblock Anchors: High-precision model-agnostic explanations.
\newblock In Sheila~A. McIlraith and Kilian~Q. Weinberger, editors, {\em
  Proceedings of the Thirty-Second {AAAI} Conference on Artificial
  Intelligence, (AAAI-18), the 30th innovative Applications of Artificial
  Intelligence (IAAI-18), and the 8th {AAAI} Symposium on Educational Advances
  in Artificial Intelligence (EAAI-18), New Orleans, Louisiana, USA, February
  2-7, 2018}, pages 1527--1535. {AAAI} Press, 2018.

\bibitem{DBLP:reference/rsh/RicciRS11}
Francesco Ricci, Lior Rokach, and Bracha Shapira.
\newblock Introduction to recommender systems handbook.
\newblock In Francesco Ricci, Lior Rokach, Bracha Shapira, and Paul~B. Kantor,
  editors, {\em Recommender Systems Handbook}, pages 1--35. Springer, 2011.

\bibitem{DBLP:conf/aaai/RossD18}
Andrew~Slavin Ross and Finale Doshi{-}Velez.
\newblock Improving the adversarial robustness and interpretability of deep
  neural networks by regularizing their input gradients.
\newblock In Sheila~A. McIlraith and Kilian~Q. Weinberger, editors, {\em
  Proceedings of the Thirty-Second {AAAI} Conference on Artificial
  Intelligence, (AAAI-18), the 30th innovative Applications of Artificial
  Intelligence (IAAI-18), and the 8th {AAAI} Symposium on Educational Advances
  in Artificial Intelligence (EAAI-18), New Orleans, Louisiana, USA, February
  2-7, 2018}, pages 1660--1669. {AAAI} Press, 2018.

\bibitem{rudin2019stop}
Cynthia Rudin.
\newblock Stop explaining black box machine learning models for high stakes
  decisions and use interpretable models instead.
\newblock {\em Nature Machine Intelligence}, 2019.

\bibitem{DBLP:conf/nips/SabourFH17}
Sara Sabour, Nicholas Frosst, and Geoffrey~E. Hinton.
\newblock Dynamic routing between capsules.
\newblock In Isabelle Guyon, Ulrike von Luxburg, Samy Bengio, Hanna~M. Wallach,
  Rob Fergus, S.~V.~N. Vishwanathan, and Roman Garnett, editors, {\em Advances
  in Neural Information Processing Systems 30: Annual Conference on Neural
  Information Processing Systems 2017, December 4-9, 2017, Long Beach, CA,
  {USA}}, pages 3856--3866, 2017.

\bibitem{DBLP:journals/tnn/SamekBMLM17}
Wojciech Samek, Alexander Binder, Gr{\'{e}}goire Montavon, Sebastian
  Lapuschkin, and Klaus{-}Robert M{\"{u}}ller.
\newblock Evaluating the visualization of what a deep neural network has
  learned.
\newblock {\em {IEEE} Trans. Neural Networks Learn. Syst.}, 28(11):2660--2673,
  2017.

\bibitem{DBLP:journals/pieee/SamekMLAM21}
Wojciech Samek, Gr{\'{e}}goire Montavon, Sebastian Lapuschkin, Christopher~J.
  Anders, and Klaus{-}Robert M{\"{u}}ller.
\newblock Explaining deep neural networks and beyond: {A} review of methods and
  applications.
\newblock {\em Proc. {IEEE}}, 109(3):247--278, 2021.

\bibitem{DBLP:journals/ijcv/SelvarajuCDVPB20}
Ramprasaath~R. Selvaraju, Michael Cogswell, Abhishek Das, Ramakrishna Vedantam,
  Devi Parikh, and Dhruv Batra.
\newblock {Grad-CAM}: Visual explanations from deep networks via gradient-based
  localization.
\newblock {\em Int. J. Comput. Vis.}, 128(2):336--359, 2020.

\bibitem{DBLP:journals/artmed/SenguptaSLGL20}
Sourya Sengupta, Amitojdeep Singh, Henry~A. Leopold, Tanmay Gulati, and
  Vasudevan Lakshminarayanan.
\newblock Ophthalmic diagnosis using deep learning with fundus images - {A}
  critical review.
\newblock {\em Artif. Intell. Medicine}, 102:101758, 2020.

\bibitem{DBLP:conf/recsys/SeoHYL17}
Sungyong Seo, Jing Huang, Hao Yang, and Yan Liu.
\newblock Interpretable convolutional neural networks with dual local and
  global attention for review rating prediction.
\newblock In Paolo Cremonesi, Francesco Ricci, Shlomo Berkovsky, and Alexander
  Tuzhilin, editors, {\em Proceedings of the Eleventh {ACM} Conference on
  Recommender Systems, RecSys 2017, Como, Italy, August 27-31, 2017}, pages
  297--305. {ACM}, 2017.

\bibitem{DBLP:conf/acl/SerranoS19}
Sofia Serrano and Noah~A. Smith.
\newblock Is attention interpretable?
\newblock In Anna Korhonen, David~R. Traum, and Llu{\'{\i}}s M{\`{a}}rquez,
  editors, {\em Proceedings of the 57th Conference of the Association for
  Computational Linguistics, {ACL}}. Association for Computational Linguistics,
  2019.

\bibitem{DBLP:conf/cvpr/ShenZ21}
Yujun Shen and Bolei Zhou.
\newblock Closed-form factorization of latent semantics in gans.
\newblock In {\em {IEEE} Conference on Computer Vision and Pattern Recognition,
  {CVPR} 2021, virtual, June 19-25, 2021}, pages 1532--1540. Computer Vision
  Foundation / {IEEE}, 2021.

\bibitem{DBLP:conf/icml/ShrikumarGK17}
Avanti Shrikumar, Peyton Greenside, and Anshul Kundaje.
\newblock Learning important features through propagating activation
  differences.
\newblock In Doina Precup and Yee~Whye Teh, editors, {\em Proceedings of the
  34th International Conference on Machine Learning, {ICML} 2017, Sydney, NSW,
  Australia, 6-11 August 2017}, volume~70 of {\em Proceedings of Machine
  Learning Research}, pages 3145--3153. {PMLR}, 2017.

\bibitem{DBLP:journals/nature/SilverHMGSDSAPL16}
David Silver, Aja Huang, Chris~J. Maddison, Arthur Guez, Laurent Sifre, George
  van~den Driessche, Julian Schrittwieser, Ioannis Antonoglou, Vedavyas
  Panneershelvam, Marc Lanctot, Sander Dieleman, Dominik Grewe, John Nham, Nal
  Kalchbrenner, Ilya Sutskever, Timothy~P. Lillicrap, Madeleine Leach, Koray
  Kavukcuoglu, Thore Graepel, and Demis Hassabis.
\newblock Mastering the game of go with deep neural networks and tree search.
\newblock {\em Nat.}, 529(7587):484--489, 2016.

\bibitem{DBLP:journals/nature/SilverSSAHGHBLB17}
David Silver, Julian Schrittwieser, Karen Simonyan, Ioannis Antonoglou, Aja
  Huang, Arthur Guez, Thomas Hubert, Lucas Baker, Matthew Lai, Adrian Bolton,
  Yutian Chen, Timothy~P. Lillicrap, Fan Hui, Laurent Sifre, George van~den
  Driessche, Thore Graepel, and Demis Hassabis.
\newblock Mastering the game of go without human knowledge.
\newblock {\em Nat.}, 550(7676):354--359, 2017.

\bibitem{DBLP:journals/corr/SimonyanVZ13}
Karen Simonyan, Andrea Vedaldi, and Andrew Zisserman.
\newblock Deep inside convolutional networks: Visualising image classification
  models and saliency maps.
\newblock In Yoshua Bengio and Yann LeCun, editors, {\em 2nd International
  Conference on Learning Representations, {ICLR} 2014, Banff, AB, Canada, April
  14-16, 2014, Workshop Track Proceedings}, 2014.

\bibitem{DBLP:journals/jimaging/SinghSL20}
Amitojdeep Singh, Sourya Sengupta, and Vasudevan Lakshminarayanan.
\newblock Explainable deep learning models in medical image analysis.
\newblock {\em J. Imaging}, 6(6):52, 2020.

\bibitem{DBLP:journals/corr/SmilkovTKVW17}
Daniel Smilkov, Nikhil Thorat, Been Kim, Fernanda~B. Vi{\'{e}}gas, and Martin
  Wattenberg.
\newblock Smoothgrad: removing noise by adding noise.
\newblock {\em CoRR}, abs/1706.03825, 2017.

\bibitem{DBLP:conf/nips/SrinivasF19}
Suraj Srinivas and Fran{\c{c}}ois Fleuret.
\newblock Full-gradient representation for neural network visualization.
\newblock In Hanna~M. Wallach, Hugo Larochelle, Alina Beygelzimer, Florence
  d'Alch{\'{e}}{-}Buc, Emily~B. Fox, and Roman Garnett, editors, {\em Advances
  in Neural Information Processing Systems 32: Annual Conference on Neural
  Information Processing Systems 2019, NeurIPS 2019, December 8-14, 2019,
  Vancouver, BC, Canada}, pages 4126--4135, 2019.

\bibitem{DBLP:journals/tvcg/StrobeltGBPPR19}
Hendrik Strobelt, Sebastian Gehrmann, Michael Behrisch, Adam Perer, Hanspeter
  Pfister, and Alexander~M. Rush.
\newblock Seq2seq-vis: {A} visual debugging tool for sequence-to-sequence
  models.
\newblock {\em {IEEE} Trans. Vis. Comput. Graph.}, 2019.

\bibitem{DBLP:journals/tvcg/StrobeltGPR18}
Hendrik Strobelt, Sebastian Gehrmann, Hanspeter Pfister, and Alexander~M. Rush.
\newblock Lstmvis: {A} tool for visual analysis of hidden state dynamics in
  recurrent neural networks.
\newblock {\em {IEEE} Trans. Vis. Comput. Graph.}, 2018.

\bibitem{DBLP:journals/corr/abs-1904-09223}
Yu~Sun, Shuohuan Wang, Yu{-}Kun Li, Shikun Feng, Xuyi Chen, Han Zhang, Xin
  Tian, Danxiang Zhu, Hao Tian, and Hua Wu.
\newblock {ERNIE:} enhanced representation through knowledge integration.
\newblock {\em CoRR}, abs/1904.09223, 2019.

\bibitem{DBLP:conf/icml/SundararajanTY17}
Mukund Sundararajan, Ankur Taly, and Qiqi Yan.
\newblock Axiomatic attribution for deep networks.
\newblock In Doina Precup and Yee~Whye Teh, editors, {\em Proceedings of the
  34th International Conference on Machine Learning, {ICML} 2017, Sydney, NSW,
  Australia, 6-11 August 2017}, volume~70 of {\em Proceedings of Machine
  Learning Research}, pages 3319--3328. {PMLR}, 2017.

\bibitem{DBLP:conf/emnlp/SwayamdiptaSLWH20}
Swabha Swayamdipta, Roy Schwartz, Nicholas Lourie, Yizhong Wang, Hannaneh
  Hajishirzi, Noah~A. Smith, and Yejin Choi.
\newblock Dataset cartography: Mapping and diagnosing datasets with training
  dynamics.
\newblock In Bonnie Webber, Trevor Cohn, Yulan He, and Yang Liu, editors, {\em
  Proceedings of the 2020 Conference on Empirical Methods in Natural Language
  Processing, {EMNLP} 2020, Online, November 16-20, 2020}, pages 9275--9293.
  Association for Computational Linguistics, 2020.

\bibitem{DBLP:conf/wsdm/TangW18}
Jiaxi Tang and Ke~Wang.
\newblock Personalized top-n sequential recommendation via convolutional
  sequence embedding.
\newblock In Yi~Chang, Chengxiang Zhai, Yan Liu, and Yoelle Maarek, editors,
  {\em Proceedings of the Eleventh {ACM} International Conference on Web Search
  and Data Mining, {WSDM} 2018, Marina Del Rey, CA, USA, February 5-9, 2018},
  pages 565--573. {ACM}, 2018.

\bibitem{DBLP:journals/tnn/TjoaG21}
Erico Tjoa and Cuntai Guan.
\newblock A survey on explainable artificial intelligence {(XAI):} toward
  medical {XAI}.
\newblock {\em {IEEE} Trans. Neural Networks Learn. Syst.}, 32(11):4793--4813,
  2021.

\bibitem{DBLP:conf/iclr/TonevaSCTBG19}
Mariya Toneva, Alessandro Sordoni, Remi~Tachet des Combes, Adam Trischler,
  Yoshua Bengio, and Geoffrey~J. Gordon.
\newblock An empirical study of example forgetting during deep neural network
  learning.
\newblock In {\em 7th International Conference on Learning Representations,
  {ICLR} 2019, New Orleans, LA, USA, May 6-9, 2019}. OpenReview.net, 2019.

\bibitem{DBLP:conf/iclr/TsiprasSETM19}
Dimitris Tsipras, Shibani Santurkar, Logan Engstrom, Alexander Turner, and
  Aleksander Madry.
\newblock Robustness may be at odds with accuracy.
\newblock In {\em 7th International Conference on Learning Representations,
  {ICLR} 2019, New Orleans, LA, USA, May 6-9, 2019}. OpenReview.net, 2019.

\bibitem{DBLP:journals/corr/abs-1907-03039}
Ilse van~der Linden, Hinda Haned, and Evangelos Kanoulas.
\newblock Global aggregations of local explanations for black box models.
\newblock {\em CoRR}, abs/1907.03039, 2019.

\bibitem{DBLP:journals/corr/abs-2010-10596}
Sahil Verma, John~P. Dickerson, and Keegan Hines.
\newblock Counterfactual explanations for machine learning: {A} review.
\newblock {\em CoRR}, abs/2010.10596, 2020.

\bibitem{DBLP:journals/nature/VinyalsBCMDCCPE19}
Oriol Vinyals, Igor Babuschkin, Wojciech~M. Czarnecki, Micha{\"{e}}l Mathieu,
  Andrew Dudzik, Junyoung Chung, David~H. Choi, Richard Powell, Timo Ewalds,
  Petko Georgiev, Junhyuk Oh, Dan Horgan, Manuel Kroiss, Ivo Danihelka, Aja
  Huang, Laurent Sifre, Trevor Cai, John~P. Agapiou, Max Jaderberg,
  Alexander~Sasha Vezhnevets, R{\'{e}}mi Leblond, Tobias Pohlen, Valentin
  Dalibard, David Budden, Yury Sulsky, James Molloy, Tom~Le Paine, {\c{C}}aglar
  G{\"{u}}l{\c{c}}ehre, Ziyu Wang, Tobias Pfaff, Yuhuai Wu, Roman Ring, Dani
  Yogatama, Dario W{\"{u}}nsch, Katrina McKinney, Oliver Smith, Tom Schaul,
  Timothy~P. Lillicrap, Koray Kavukcuoglu, Demis Hassabis, Chris Apps, and
  David Silver.
\newblock Grandmaster level in starcraft {II} using multi-agent reinforcement
  learning.
\newblock {\em Nat.}, 575(7782):350--354, 2019.

\bibitem{DBLP:conf/acl/VoitaTMST19}
Elena Voita, David Talbot, Fedor Moiseev, Rico Sennrich, and Ivan Titov.
\newblock Analyzing multi-head self-attention: Specialized heads do the heavy
  lifting, the rest can be pruned.
\newblock In Anna Korhonen, David~R. Traum, and Llu{\'{\i}}s M{\`{a}}rquez,
  editors, {\em Proceedings of the 57th Conference of the Association for
  Computational Linguistics, {ACL} 2019, Florence, Italy, July 28- August 2,
  2019, Volume 1: Long Papers}, pages 5797--5808. Association for Computational
  Linguistics, 2019.

\bibitem{DBLP:journals/corr/abs-1912-10920}
Andrey Voynov and Artem Babenko.
\newblock {RPGAN:} gans interpretability via random routing.
\newblock {\em CoRR}, abs/1912.10920, 2019.

\bibitem{DBLP:conf/icml/VoynovB20}
Andrey Voynov and Artem Babenko.
\newblock Unsupervised discovery of interpretable directions in the {GAN}
  latent space.
\newblock In {\em Proceedings of the 37th International Conference on Machine
  Learning, {ICML} 2020, 13-18 July 2020, Virtual Event}, volume 119 of {\em
  Proceedings of Machine Learning Research}, pages 9786--9796. {PMLR}, 2020.

\bibitem{DBLP:journals/corr/abs-1906-02032}
Minh~N. Vu, Truc D.~T. Nguyen, NhatHai Phan, Ralucca Gera, and My~T. Thai.
\newblock Evaluating explainers via perturbation.
\newblock {\em CoRR}, abs/1906.02032, 2019.

\bibitem{DBLP:journals/corr/abs-1711-00399}
Sandra Wachter, Brent~D. Mittelstadt, and Chris Russell.
\newblock Counterfactual explanations without opening the black box: Automated
  decisions and the {GDPR}.
\newblock {\em CoRR}, abs/1711.00399, 2017.

\bibitem{DBLP:conf/cvpr/WangWDYZDMH20}
Haofan Wang, Zifan Wang, Mengnan Du, Fan Yang, Zijian Zhang, Sirui Ding, Piotr
  Mardziel, and Xia Hu.
\newblock {Score-CAM}: Score-weighted visual explanations for convolutional
  neural networks.
\newblock In {\em 2020 {IEEE/CVF} Conference on Computer Vision and Pattern
  Recognition, {CVPR} Workshops 2020, Seattle, WA, USA, June 14-19, 2020},
  pages 111--119. Computer Vision Foundation / {IEEE}, 2020.

\bibitem{welinderetal2010caltech}
P.~Welinder, S.~Branson, T.~Mita, C.~Wah, F.~Schroff, S.~Belongie, and
  P.~Perona.
\newblock {Caltech-UCSD birds 200}.
\newblock Technical Report CNS-TR-2010-001, California Institute of Technology,
  2010.

\bibitem{DBLP:conf/nips/WickramanayakeH21}
Sandareka Wickramanayake, Wynne Hsu, and Mong{-}Li Lee.
\newblock Explanation-based data augmentation for image classification.
\newblock In Marc'Aurelio Ranzato, Alina Beygelzimer, Yann~N. Dauphin, Percy
  Liang, and Jennifer~Wortman Vaughan, editors, {\em Advances in Neural
  Information Processing Systems 34: Annual Conference on Neural Information
  Processing Systems 2021, NeurIPS 2021, December 6-14, 2021, virtual}, pages
  20929--20940, 2021.

\bibitem{DBLP:conf/emnlp/WiegreffeP19}
Sarah Wiegreffe and Yuval Pinter.
\newblock Attention is not not explanation.
\newblock In Kentaro Inui, Jing Jiang, Vincent Ng, and Xiaojun Wan, editors,
  {\em Proceedings of the 2019 Conference on Empirical Methods in Natural
  Language Processing and the 9th International Joint Conference on Natural
  Language Processing, {EMNLP-IJCNLP}}. Association for Computational
  Linguistics, 2019.

\bibitem{DBLP:conf/eccv/WooPLK18}
Sanghyun Woo, Jongchan Park, Joon{-}Young Lee, and In~So Kweon.
\newblock {CBAM:} convolutional block attention module.
\newblock In Vittorio Ferrari, Martial Hebert, Cristian Sminchisescu, and Yair
  Weiss, editors, {\em Computer Vision - {ECCV} 2018 - 15th European
  Conference, Munich, Germany, September 8-14, 2018, Proceedings, Part {VII}},
  volume 11211 of {\em Lecture Notes in Computer Science}, pages 3--19.
  Springer, 2018.

\bibitem{DBLP:journals/corr/abs-2006-16789}
Guandong Xu, Tri~Dung Duong, Qian Li, Shaowu Liu, and Xianzhi Wang.
\newblock Causality learning: {A} new perspective for interpretable machine
  learning.
\newblock {\em CoRR}, abs/2006.16789, 2020.

\bibitem{DBLP:journals/ijcv/YangSZ21}
Ceyuan Yang, Yujun Shen, and Bolei Zhou.
\newblock Semantic hierarchy emerges in deep generative representations for
  scene synthesis.
\newblock {\em Int. J. Comput. Vis.}, 129(5):1451--1466, 2021.

\bibitem{yang2019benchmarking}
Mengjiao Yang and Been Kim.
\newblock Benchmarking attribution methods with relative feature importance.
\newblock {\em arXiv}, 2019.

\bibitem{DBLP:conf/cvpr/Yao0XZS0T021}
Yazhou Yao, Tao Chen, Guo{-}Sen Xie, Chuanyi Zhang, Fumin Shen, Qi~Wu, Zhenmin
  Tang, and Jian Zhang.
\newblock Non-salient region object mining for weakly supervised semantic
  segmentation.
\newblock In {\em {IEEE} Conference on Computer Vision and Pattern Recognition,
  {CVPR} 2021, virtual, June 19-25, 2021}, pages 2623--2632. Computer Vision
  Foundation / {IEEE}, 2021.

\bibitem{DBLP:conf/nips/YehHSIR19}
Chih{-}Kuan Yeh, Cheng{-}Yu Hsieh, Arun~Sai Suggala, David~I. Inouye, and
  Pradeep Ravikumar.
\newblock On the (in)fidelity and sensitivity of explanations.
\newblock In Hanna~M. Wallach, Hugo Larochelle, Alina Beygelzimer, Florence
  d'Alch{\'{e}}{-}Buc, Emily~B. Fox, and Roman Garnett, editors, {\em Advances
  in Neural Information Processing Systems 32: Annual Conference on Neural
  Information Processing Systems 2019, NeurIPS 2019, December 8-14, 2019,
  Vancouver, BC, Canada}, pages 10965--10976, 2019.

\bibitem{DBLP:conf/nips/YingBYZL19}
Zhitao Ying, Dylan Bourgeois, Jiaxuan You, Marinka Zitnik, and Jure Leskovec.
\newblock Gnnexplainer: Generating explanations for graph neural networks.
\newblock In Hanna~M. Wallach, Hugo Larochelle, Alina Beygelzimer, Florence
  d'Alch{\'{e}}{-}Buc, Emily~B. Fox, and Roman Garnett, editors, {\em Advances
  in Neural Information Processing Systems 32: Annual Conference on Neural
  Information Processing Systems 2019, NeurIPS 2019, December 8-14, 2019,
  Vancouver, BC, Canada}, pages 9240--9251, 2019.

\bibitem{yuan2021explaining}
Tingyi Yuan, Xuhong Li, Haoyi Xiong, Hui Cao, and Dejing Dou.
\newblock Explaining information flow inside vision transformers using markov
  chain.
\newblock In {\em Neural Information Processing Systems XAI4Debugging
  Workshop}, 2021.

\bibitem{DBLP:conf/iclr/ZagoruykoK17}
Sergey Zagoruyko and Nikos Komodakis.
\newblock Paying more attention to attention: Improving the performance of
  convolutional neural networks via attention transfer.
\newblock In {\em 5th International Conference on Learning Representations,
  {ICLR} 2017, Toulon, France, April 24-26, 2017, Conference Track
  Proceedings}. OpenReview.net, 2017.

\bibitem{DBLP:journals/cacm/ZhangBHRV21}
Chiyuan Zhang, Samy Bengio, Moritz Hardt, Benjamin Recht, and Oriol Vinyals.
\newblock Understanding deep learning (still) requires rethinking
  generalization.
\newblock {\em Commun. {ACM}}, 64(3):107--115, 2021.

\bibitem{DBLP:conf/iclr/ZhangCDL18}
Hongyi Zhang, Moustapha Ciss{\'{e}}, Yann~N. Dauphin, and David Lopez{-}Paz.
\newblock Mixup: Beyond empirical risk minimization.
\newblock In {\em 6th International Conference on Learning Representations,
  {ICLR} 2018, Vancouver, BC, Canada, April 30 - May 3, 2018, Conference Track
  Proceedings}. OpenReview.net, 2018.

\bibitem{DBLP:journals/ijcv/ZhangBLBSS18}
Jianming Zhang, Sarah~Adel Bargal, Zhe Lin, Jonathan Brandt, Xiaohui Shen, and
  Stan Sclaroff.
\newblock Top-down neural attention by excitation backprop.
\newblock {\em Int. J. Comput. Vis.}, 126(10):1084--1102, 2018.

\bibitem{DBLP:conf/aaai/ZhangCSWZ18}
Quanshi Zhang, Ruiming Cao, Feng Shi, Ying~Nian Wu, and Song{-}Chun Zhu.
\newblock Interpreting {CNN} knowledge via an explanatory graph.
\newblock In Sheila~A. McIlraith and Kilian~Q. Weinberger, editors, {\em
  Proceedings of the Thirty-Second {AAAI} Conference on Artificial
  Intelligence, (AAAI-18), the 30th innovative Applications of Artificial
  Intelligence (IAAI-18), and the 8th {AAAI} Symposium on Educational Advances
  in Artificial Intelligence (EAAI-18), New Orleans, Louisiana, USA, February
  2-7, 2018}, pages 4454--4463. {AAAI} Press, 2018.

\bibitem{DBLP:conf/cvpr/ZhangWZ18a}
Quanshi Zhang, Ying~Nian Wu, and Song{-}Chun Zhu.
\newblock Interpretable convolutional neural networks.
\newblock In {\em 2018 {IEEE} Conference on Computer Vision and Pattern
  Recognition, {CVPR} 2018, Salt Lake City, UT, USA, June 18-22, 2018}, pages
  8827--8836. Computer Vision Foundation / {IEEE} Computer Society, 2018.

\bibitem{DBLP:conf/cvpr/ZhangYMW19}
Quanshi Zhang, Yu~Yang, Haotian Ma, and Ying~Nian Wu.
\newblock Interpreting cnns via decision trees.
\newblock In {\em {IEEE} Conference on Computer Vision and Pattern Recognition,
  {CVPR} 2019, Long Beach, CA, USA, June 16-20, 2019}, pages 6261--6270.
  Computer Vision Foundation / {IEEE}, 2019.

\bibitem{DBLP:journals/csur/ZhangYST19}
Shuai Zhang, Lina Yao, Aixin Sun, and Yi~Tay.
\newblock Deep learning based recommender system: {A} survey and new
  perspectives.
\newblock {\em {ACM} Comput. Surv.}, 52(1):5:1--5:38, 2019.

\bibitem{DBLP:conf/icml/ZhangZ19}
Tianyuan Zhang and Zhanxing Zhu.
\newblock Interpreting adversarially trained convolutional neural networks.
\newblock In Kamalika Chaudhuri and Ruslan Salakhutdinov, editors, {\em
  Proceedings of the 36th International Conference on Machine Learning, {ICML}
  2019, 9-15 June 2019, Long Beach, California, {USA}}, volume~97 of {\em
  Proceedings of Machine Learning Research}, pages 7502--7511. {PMLR}, 2019.

\bibitem{DBLP:journals/ftir/ZhangC20}
Yongfeng Zhang and Xu~Chen.
\newblock Explainable recommendation: {A} survey and new perspectives.
\newblock {\em Found. Trends Inf. Retr.}, 14(1):1--101, 2020.

\bibitem{DBLP:conf/miccai/ZhaoZWJX18}
Guannan Zhao, Bo~Zhou, Kaiwen Wang, Rui Jiang, and Min Xu.
\newblock {Respond-CAM}: Analyzing deep models for 3d imaging data by
  visualizations.
\newblock In Alejandro~F. Frangi, Julia~A. Schnabel, Christos Davatzikos,
  Carlos Alberola{-}L{\'{o}}pez, and Gabor Fichtinger, editors, {\em Medical
  Image Computing and Computer Assisted Intervention - {MICCAI} 2018 - 21st
  International Conference, Granada, Spain, September 16-20, 2018, Proceedings,
  Part {I}}, volume 11070 of {\em Lecture Notes in Computer Science}, pages
  485--492. Springer, 2018.

\bibitem{DBLP:conf/cvpr/ZhouKLOT16}
Bolei Zhou, Aditya Khosla, {\`{A}}gata Lapedriza, Aude Oliva, and Antonio
  Torralba.
\newblock Learning deep features for discriminative localization.
\newblock In {\em 2016 {IEEE} Conference on Computer Vision and Pattern
  Recognition, {CVPR} 2016, Las Vegas, NV, USA, June 27-30, 2016}, pages
  2921--2929. {IEEE} Computer Society, 2016.

\end{thebibliography}
}
    
\end{document}